\pdfoutput=1
\documentclass[11pt]{article}

\usepackage{acl}

\usepackage{times}
\usepackage{latexsym}

\usepackage[T1]{fontenc}
\usepackage[utf8]{inputenc}

\usepackage{microtype}

\usepackage{inconsolata}

\usepackage{graphicx}

\usepackage{booktabs} % 表格边框
\usepackage{array}
\usepackage{stfloats}  % 强制双栏图片在本页
\usepackage{arydshln}
\usepackage{ulem} % 删除线
 
\usepackage{enumitem}  % 调整列表间距
\usepackage{xcolor}
\usepackage{titlesec}
\titlespacing*{\paragraph}{0pt}{0.15\baselineskip}{\baselineskip}

\title{Is LLM an Overconfident Judge? Unveiling the Capabilities of LLMs in Detecting Offensive Language with Annotation Disagreement
}

\author{Junyu Lu\textsuperscript{1}, Kai Ma\textsuperscript{1}, Kaichun Wang\textsuperscript{1}, Kelaiti Xiao\textsuperscript{1,2}\\
\textbf{Roy Ka-Wei Lee\textsuperscript{3}, Bo Xu\textsuperscript{1}, Liang Yang\textsuperscript{1}, Hongfei Lin\textsuperscript{1}\thanks{Corresponding author.}} \\
        \textsuperscript{1}Key Laboratory of Social Computing and Cognitive Intelligence, Dalian University of Technology \\ 
        \textsuperscript{2}School of Computer Science and Technology, Xinjiang Normal University\\
        \textsuperscript{3}Social AI Studio, Singapore University of Technology and Design \\ 
        \texttt{dutljy@mail.dlut.edu.cn, hflin@dlut.edu.cn}}

\begin{document}
\maketitle
\begin{abstract}

Large Language Models (LLMs) have become essential for offensive language detection, yet their ability to handle annotation disagreement remains underexplored. Disagreement samples, which arise from subjective interpretations, pose a unique challenge due to their ambiguous nature. Understanding how LLMs process these cases, particularly their confidence levels, can offer insight into their alignment with human annotators. This study systematically evaluates the performance of multiple LLMs in detecting offensive language at varying levels of annotation agreement. We analyze binary classification accuracy, examine the relationship between model confidence and human disagreement, and explore how disagreement samples influence model decision-making during few-shot learning and instruction fine-tuning. Our findings reveal that LLMs struggle with low-agreement samples, often exhibiting overconfidence in these ambiguous cases. However, utilizing disagreement samples in training improves both detection accuracy and model alignment with human judgment. These insights provide a foundation for enhancing LLM-based offensive language detection in real-world moderation tasks.

\textcolor{red}{\textit{Disclaimer}:\textit{ The paper contains content that may be profane, vulgar, or offensive.}}
% \textit{\textbf{Disclaimer}: The samples presented by this paper may be considered offensive or vulgar.}

\end{abstract}

\section{Introduction}

\begin{table*}[h]
\small
  \centering
    \begin{tabular}{m{2.5cm}<{\centering}m{8.5cm}m{1.5cm}<{\centering}m{1.5cm}<{\centering}}
    \toprule
    Agreement Degree & Example & Hard Label & Soft Label \\
    \midrule
    $N^{++}$   & Stand for something or else fall for anything. \#BlackLivesMatter Hello world! What a great day to be alive \#Trump2020 \#MAGA & 0     & 0 \\
    \midrule
    $N^{+}$   & Come on man! Lock’em up!!! \#maga Not the first time. You all misspelled \#blacklivesmatter. Speak up! @user & 0     & 0.2 \\
    \midrule
    $N^{0}$    & Set fire to Fox News (metaphorically) @user is outing \#BLACK-LIVES-MATTER as a cult! HE IS CORRECT! & 0     & 0.4 \\
    \midrule
    $O^{0}$   & \#DISGUSTING \#Democrats terrorize old folks just before \#elections2020 I love this shit! \#BlackLivesMatter & 1     & 0.6 \\
    \midrule
    $O^{+}$    & @user You’re a bumbling fool \#elections2020 Elections 2020: Red Rapist v. Blue Racist & 1     & 0.8 \\
    \midrule
    $O^{++}$   & Y’all trending about kpop stans instead of \#BlackLivesMatter big fack you Crazy idiots. This is batshit bullshit. \#elections2020 & 1     & 1 \\
    \bottomrule
    \end{tabular}%
  \vspace{-0.025in}
  \caption{Eamples with different degrees of annotation agreement, where $N$ denotes non-offensive language and $O$ denotes offensive language. The superscripts ++, +, and 0 represent unanimous, medium, and low agreement, respectively. These samples are collected from the MD-Agreement dataset \cite{DBLP:conf/emnlp/LeonardelliMAGT21}.}
  \vspace{-0.075in}
  \label{examples}%
\end{table*}%

\paragraph{Motivation.} A fundamental challenge in offensive language detection is annotation disagreement—cases where human annotators provide conflicting labels for the same text. Disagreement arises due to differences in individual perception, cultural context, and linguistic ambiguity, making offensive language detection inherently subjective \cite{DBLP:conf/www/AroyoDTRR19, DBLP:conf/aiia/Basile20, DBLP:journals/jair/UmaFHPPP21}. However, prior research predominantly treats this task as a binary classification problem, assuming consensus among annotators and failing to account for the inherent subjectivity in offensive language perception.

While large language models (LLMs) have been extensively applied to offensive language detection \cite{DBLP:conf/icwsm/0006AD24, DBLP:conf/www/HuangKA23a}, existing studies primarily evaluate their performance on datasets with binary labels, overlooking their ability to handle cases where annotators disagree. This oversimplification limits our understanding of how well LLMs align with human judgment in ambiguous cases. Moreover, models may exhibit overconfidence in cases where human annotators themselves are uncertain, raising concerns about their reliability for real-world moderation.

\paragraph{Research Objectives.} To bridge this gap, we systematically evaluate LLMs’ ability to process disagreement samples, analyzing both classification accuracy and model confidence. Through this, we seek to determine whether LLMs can effectively navigate subjective offensive language judgments and align with human reasoning. An ideal model should express high confidence for unanimously labeled cases and lower confidence for ambiguous samples, reflecting their inherent uncertainty \cite{weerasooriya-etal-2023-disagreement, baumler-etal-2023-examples, uma-etal-2021-semeval, DBLP:conf/semeval/LeonardelliAABF23}. Our study provides insights into whether LLMs capture these nuances or exhibit overconfidence in disagreement cases, which could undermine their trustworthiness in content moderation.

This paper systematically investigates how LLMs handle annotation disagreement in offensive language detection. Specifically, we address the following research questions: (\textbf{RQ1}) To what extent can LLMs accurately detect offensive language in cases of human annotation disagreement? (\textbf{RQ2}) How do disagreement samples shape LLM learning and influence decision-making?

To answer RQ1, we evaluate multiple open-source and closed-source LLMs in a zero-shot setting, analyzing both classification accuracy and the relationship between annotation agreement and model confidence. For RQ2, we examine the impact of disagreement samples in few-shot learning and instruction fine-tuning, assessing how different agreement levels affect model performance.

\paragraph{Contributions.} We summarize the contributions of this paper as follows:
\begin{itemize}[itemsep=0.125pt]
    \item We provide the first systematic evaluation of LLMs' performance in offensive language detection under annotation disagreement, revealing key insights into model reliability and human-AI alignment.
    \item We conduct an extensive empirical study on LLMs' handling of disagreement cases, examining models' performance, confidence, and alignment with human judgment.
    \item We analyze the impact of training on disagreement samples, demonstrating how few-shot learning and instruction fine-tuning on these samples influence LLM decision-making in offensive language detection.
\end{itemize}

\section{Preliminary}

% In this study, we evaluate the binary classification performance of various LLMs on detecting offensive language, focus on samples with varying degrees of annotation agreement. 
% Additionally, we estimate the relationship between the degree of annotator agreement and the LLM confidence.
% This includes an overview of the dataset, the specific models evaluated, and the key experimental settings.

\subsection{Dataset}

Since annotation disagreements can stem from both intrinsic linguistic ambiguity and labeling error, selecting an appropriate benchmark dataset requires meeting two key criteria: (1) high annotation quality to ensure reliability, and (2) open access to unaggregated annotations to facilitate fine-grained analysis. To ensure a robust evaluation, we employ the MD-Agreement dataset \cite{DBLP:conf/emnlp/LeonardelliMAGT21}, a high-quality corpus for offensive language detection. It contains 10,753 tweets, each labeled by five trained human annotators, ensuring a reliable annotation process. 

The dataset provides both hard labels (majority-voted labels) and soft labels, which indicate the level of agreement among annotators. The soft labels are categorized into three levels:
% \red{(Roy: There is a misalignment in the Table 1. Soft Label is a number (e.g., 0.2) instead of this ++ labels. You might want to change Table 1 or define how the score is derived.)}
\begin{itemize}[itemsep=0.125pt]
    \item \textit{Unanimous agreement} ($A^{++}$): All five annotators agree on the label.
    \item \textit{Mild agreement} ($A^{+}$): Four out of five annotators agree.
    \item \textit{Weak agreement} ($A^{0}$): Only three annotators agree, while two disagree.
\end{itemize}

Each sample in the dataset is also classified as either \textit{offensive} or \textit{non-offensive}, following the same agreement-level framework:
\begin{itemize}[itemsep=0.125pt]
    \item \textbf{Offensive samples} ($O^{++}$, $O^{+}$, $O^{0}$): Instances labeled as offensive, where the agreement level corresponds to the unanimous, mild, or weak agreement, respectively.
    \item \textbf{Non-offensive samples} ($N^{++}$, $N^{+}$, $N^{0}$): Instances labeled as non-offensive, with the same agreement-level distinctions.
\end{itemize}

Thus, the agreement notation ($++$, $+$, $0$) applies uniformly across both offensive and non-offensive categories, ensuring consistency in the dataset's annotation schema. To facilitate subsequent research, we convert the soft labels into floating-point numbers in the range [0, 1] by averaging the hard labels from five annotators (offensive as 1, non-offensive as 0) for each sample.

Examples of samples across different agreement levels are provided in Table \ref{examples}, and the overall dataset statistics are presented in Table \ref{statistics}. 
% The reliability of MD-Agreement’s annotations has been independently validated in prior studies \cite{DBLP:conf/eacl/SandriLTJ23}, confirming its suitability for benchmarking offensive language detection.
The reliability of MD-Agreement’s annotations has been independently validated in prior studies \cite{DBLP:conf/eacl/SandriLTJ23}, confirming that the disagreement samples are caused by their inherent ambiguity, rather than labeling errors.
MD-Agreement serves as the official corpus for SemEval 2023 Task 11 \cite{DBLP:conf/semeval/LeonardelliAABF23} and has been widely utilized by researchers in the field \cite{DBLP:conf/emnlp/DengZ0W0M23, DBLP:conf/naacl/MokhberianMHBML24}. Further dataset details are provided in Appendix \ref{data_details}.

\begin{table*}[t]
\small
\renewcommand{\arraystretch}{1.2}
  \centering
  \scalebox{0.95}{
    \begin{tabular}{c|ccc|cccccc|cc|c}
    \hline
    Split  & $A^{++}$   & $A^{+}$    & $A$     & $N^{++}$   & $N^+$    & $N$     & $O$     & $O^+$    & $O^{++}$   & N-Off. & Off.  & Total \\
    \hline
    Train & 2,778  & 1,930  & 1,884  & 2,303  & 1,295  & 1,032  & 852   & 635   & 475   & 4,630  & 1,962  & 6,592 \\
    Dev  & 464   & 317   & 322   & 346   & 199   & 171   & 151   & 118   & 118   & 1,103  & 387   & 1,103 \\
    Test  & 1,292  & 909   & 803   & 1,020  & 549   & 470   & 386   & 360   & 272   & 3,004  & 1,018  & 3,057 \\
    \hline
    MD-Agreement & 4,535  & 3,156  & 3,062  & 3,669  & 2,043  & 1,673  & 1,389  & 1,113  & 866   & 7,385  & 3,368  & 10,753 \\
    \hline
    \end{tabular}%
    }
  \caption{Statistics of the MD-Agreement dataset, where $N$ denotes non-offensive language and $O$ denotes offensive language. The superscripts ++, +, and 0 represent unanimous, medium, and low agreement, respectively.}
  \vspace{-0.1in}
  \label{statistics}%
\end{table*}%

\subsection{Models}
To ensure a comprehensive evaluation, we include both closed-source and open-source LLMs, covering a range of architectures and parameter sizes. For closed-source models, we evaluate widely used proprietary LLMs, including GPT-3.5, GPT-4, GPT-4o, GPT-o1, Claude-3.5, and Gemini-1.5. %Since the current GPT-o1 API does not allow temperature adjustment, we report its binary classification performance at the default temperature of 1 in Table 3. 
%\red{One minor concern: Is it premature to mention the setting of the temperature coefficient, given that the experimental settings for binary classification and confidence estimation have not yet been introduced?}
We also evaluate three families of open-source LLMs at different scales: LLaMa-3 (8B, 70B), Qwen-2.5 (7B, 72B), and Mixtral (8x7B, 8x22B). Further details on the model versions are provided in Appendix~\ref{version}. %\red{Roy: I think there is still a part where we need to discuss the model settings. Because this is asked as part of the submission checklist.}

%To ensure a comprehensive evaluation, we access a range of LLMs, including both closed-source and open-source models:
%For closed-source models, we examine several widely used LLMs, including GPT-3.5, GPT-4, GPT-4o, GPT-o1\footnote{Since the current GPT-o1 API does not allow adjustment of the temperature coefficient, we only present its binary classification performance at a temperature of 1 in Table 3, which is the default setting.
%}, Claude-3.5, and Gemini-1.5.
%For open-source models, we utilize three LLMs with different parameter sizes: LLaMa-3 8B and 70B, Qwen-2.5 7B and 72B, and Mixtral 8x7B and 8x22B.
%More information about the versions of LLMs is provided in Appendix \ref{version}.

% A more detailed comparison of the GPT-family models can be found in Appendix.

% \subsection{Experimental Settings}

\section{RQ1: Evaluating LLMs on Offensive Language with Annotation Disagreement}
In this section, we evaluate the ability of LLMs to detect offensive language in a zero-shot setting.
We focus on two key aspects: (1) binary classification accuracy, assessing how effectively models distinguish offensive from non-offensive language across varying annotation agreement levels, and (2) model confidence, analyzing whether LLMs exhibit appropriate uncertainty in ambiguous cases. 
These aspects are essential for determining whether LLMs can reliably perform offensive language detection in real-world scenarios, where human annotators often disagree.

%\section{How Do LLMs Perform in Detecting Offensive Language with Annotation Disagreement?}

% In this section, we present the method used to evaluate both the binary accuracy in detecting offensive language and the relationship between annotators and LLMs.

%In this section, we assess the performance of LLMs in a zero-shot scenario, focusing on both the binary accuracy in detecting offensive language and the relationship between annotation agreement and LLM confidence.
%We utilize all the samples from the MD-Agreement dataset for this evaluation and analyze the performance on subsets with different annotation agreements.

\subsection{Evaluation of Binary Classification Performance}
We assess binary classification performance by evaluating LLMs in a zero-shot setting without additional fine-tuning. To ensure deterministic predictions, we set the temperature coefficient of the LLMs to 0, forcing the model to select the most probable category. We use accuracy and F1 score as evaluation metrics to measure classification performance. The prompt template used for offensive language detection is provided in Appendix \ref{template}. LLM outputs are converted into hard predictions, where 1 indicates offensive and 0 indicates non-offensive. 
We utilize all the samples from MD-Agreement for a comprehensive evaluation.
The classification results are presented in Table~\ref{main}. Based on the results, we observe the following key findings:

% \paragraph{\red{\sout{(1) LLMs achieve high accuracy for $A^{++}$ samples but struggle with ambiguity.}}} 
% \red{\sout{In the zero-shot setting, LLMs accurately classify unanimously agreed-upon ($A^{++}$) samples, achieving 88.28\% accuracy for closed-source models and 86.07\% for open-source models. Notably, LLaMa3-70B now performs comparably to proprietary models. While LLMs perform well on clear-cut cases, real-world moderation involves ambiguous or disputed samples, where performance deteriorates significantly.}}

\paragraph{(1) LLMs achieve high accuracy for unanimous agreement ($A^{++}$) samples.} 
In the zero-shot setting, LLMs consistently accurately classify unanimously agreed-upon ($A^{++}$) samples, achieving 88.28\% accuracy for closed-source models and 86.07\% for open-source models. Notably, LLaMa3-70B now performs comparably to proprietary models. These results suggest that LLMs perform well on clear-cut cases, driven by their background knowledge and reasoning capabilities.

\paragraph{(2) LLM performance declines sharply for ambiguous cases.}  
As annotation agreement decreases, LLMs struggle to classify offensive language consistently. 
GPT-4o's F1 score drops from 85.24\% on $A^{++}$ samples to 74.6\% on $A^+$ and 57.06\% on $A^0$. Similarly, all models score below 65\% on $A^0$ samples. 
% \red{\sout{, indicating substantial uncertainty.} Junyu: In binary classification experiments, uncertainty may not be as obvious, more so reflecting the LLM's inability.}
This highlights LLMs' inability to resolve subjective cases in the real world, where human disagreement often stems from cultural, contextual, or linguistic nuances that models fail to capture.

% ,  moderation involves ambiguous or disputed samples, where performance deteriorates significantly.

\paragraph{(3) Larger models improve accuracy but do not resolve annotation disagreement.}  
While larger models generally perform better, their improvement shrinks for ambiguous cases. For example, LLaMa3-70B outperforms LLaMa3-8B by 10.64\% on $A^{++}$ samples but only by 3.11\% on $A^0$. Similarly, Mixtral and Qwen2.5 show no substantial gain in detecting disagreement samples despite increased parameters.  Model scaling alone does not resolve ambiguity, suggesting that larger models lack the nuanced human reasoning required to navigate subjective cases. 
Alternative training strategies, such as human-in-the-loop approaches or fine-tuning on disagreement samples, may be necessary.

\begin{table*}
\renewcommand{\arraystretch}{1.2}
\small
  \centering
    \begin{tabular}{m{2.5cm}<{\centering}|m{0.925cm}<{\centering}m{0.925cm}<{\centering}|m{0.925cm}<{\centering}m{0.925cm}<{\centering}|m{0.925cm}<{\centering}m{0.925cm}<{\centering}|m{0.925cm}<{\centering}m{0.925cm}<{\centering}}
    \hline
          & \multicolumn{2}{c|}{Overall} & \multicolumn{2}{c|}{$A^{++}$} & \multicolumn{2}{c|}{$A^{+}$} & \multicolumn{2}{c}{$A^{o}$} \\
    \hline
    Model & Acc.$\;\uparrow$   & F1$\;\uparrow$    & Acc.$\;\uparrow$   & F1$\;\uparrow$    & Acc.$\;\uparrow$   & F1$\;\uparrow$    & Acc.$\;\uparrow$   & F1$\;\uparrow$ \\
    \hline
    \multicolumn{9}{c}{\textit{Closed-Source Large Language Models \ \textit{(CS-LLMs)}}} \\
    \hline
    GPT-o1 & 78.35 & 69.03 & 91.95 & 81.29 & \underline{77.50} & \underline{72.03} & 59.08 & 58.63 \\
    GPT-4o & \textbf{80.36} & 70.33 & \textbf{93.96} & \textbf{85.24} & \textbf{80.67} & \textbf{74.60} & \underline{59.90} & 57.06 \\
    GPT-4 & 74.18 & 69.07 & 88.64 & 76.75 & 70.12 & 68.91 & 56.96 & \textbf{64.63} \\
    GPT-3.5 & 67.07 & 63.45 & 78.99 & 64.02 & 62.39 & 63.28 & 54.25 & 63.18 \\
    Claude-3.5 & \underline{78.56} & \underline{70.93} & \underline{92.59} & \underline{83.13} & 76.39 & 72.02 & \textbf{60.03} & 62.61 \\
    % Claude-3 &       &       &       &       &       &       &       &  \\
    Gemini-1.5 & 69.50  & 66.07 & 83.53 & 69.70 & 64.48 & 65.73 & 53.89 & 64.07 \\
    \hline
    \textit{Avg. of CS-LLM} & 74.67 & 68.15 & 88.28 & 76.69 & 71.93 & 69.43 & 57.35 & 61.70 \\
    \hline
    \multicolumn{9}{c}{\textit{Open-Source Large Language Models \ \textit{(OS-LLMs)}}} \\
    \hline
    LLaMa3-70B & 76.93 & \textbf{71.06} & 91.40  & 81.36 & 74.46 & 71.96 & 58.03 & \underline{64.37} \\
    LLaMa3-8B & 71.82 & 65.31 & 85.56 & 70.72 & 68.22 & 66.06 & 55.19 & 61.26 \\
    Qwen2.5-72B & 72.08 & 66.86 & 84.74 & 70.92 & 68.41 & 67.36 & 57.12 & 63.76 \\
    Qwen2.5-7B & 71.10  & 67.14 & 85.34 & 72.02 & 66.92 & 67.25 & 54.31 & 64.06 \\
    Mixtral-8x22B & 73.46 & 67.82 & 87.12 & 74.27 & 69.93 & 68.21 & 56.86 & 63.44 \\
    Mixtral-8x7B & 70.57 & 65.59 & 82.27 & 67.58 & 67.14 & 66.32 & 56.76 & 63.63 \\
    \hline
    \textit{Avg. of OS-LLM} & 72.66 & 67.30 & 86.07 & 72.81 & 69.18 & 67.86 & 56.38 & 63.42 \\
    \hline
    \end{tabular}%
  \caption{Binary classification performance of LLMs on the MD-Agreement dataset and its three subsets $A^{++}$, $A^{+}$, and $A^{0}$. \textit{Avg. of CS-LLM} and \textit{OS-LLM} respectively denote the average performance of the close-source and open-source LLMs.
  Results show the accuracy (\textit{Acc.}) and $F_1$ in percentage (\%). The \textbf{bold} and \underline{underline} scores respectively represent the optimal and suboptimal values.} 
  \vspace{-0.075in}
  \label{main}%
\end{table*}%

\begin{figure}
\centering
\includegraphics[width=7.25cm]{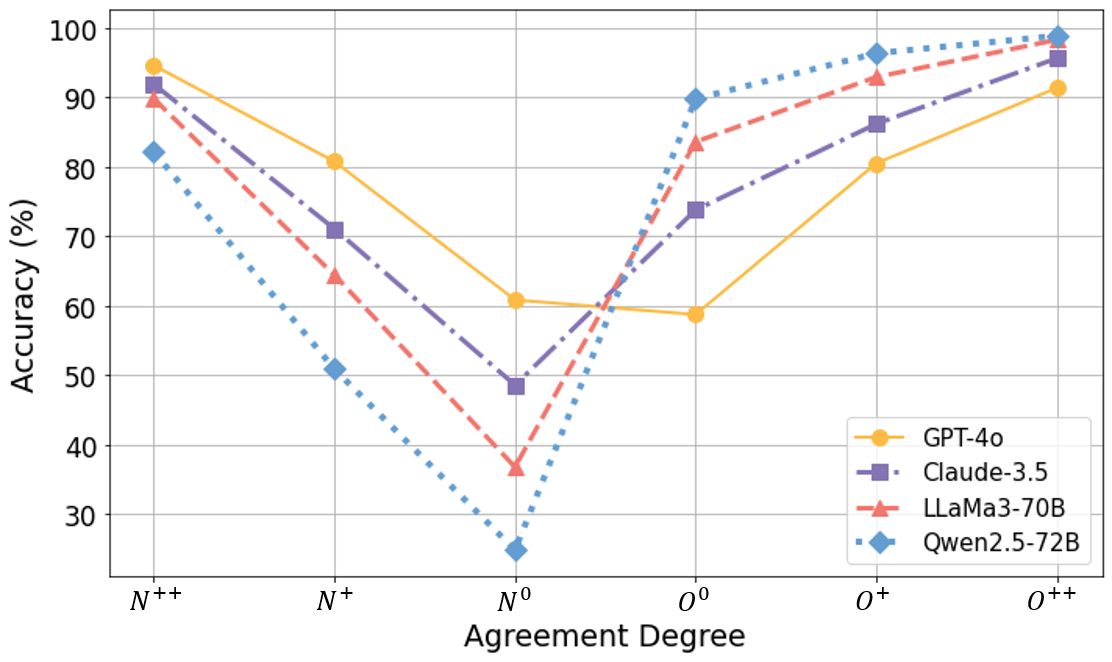}
\vspace{-0.05in}
\caption{Accuracy of LLMs on detecting offensive and non-offensive language with different degrees of annotation agreement.
}
\vspace{-0.125in}
\label{dis_type}
\end{figure}

\paragraph{(4) LLMs are biased toward classifying uncertain cases as offensive.}  
%\red{Junyu: This result is based on Figure 1, which evaluates the accuracy of several LLMs for both offensive and non-offensive language, do we need to explain this?} 
We evaluate the accuracy for offensive and non-offensive language across different agreement levels, as shown in Figure \ref{dis_type}. 
We observed that across all agreement levels, LLMs demonstrate higher accuracy in identifying offensive language than non-offensive language. 
In particular, for low-agreement non-offensive samples ($N^0$), accuracy drops to 45.77\%, indicating a strong tendency to misclassify ambiguous content as offensive. This over-sensitivity could lead to false positives in automated moderation systems, increasing the risk of justified content removal and restricting legitimate speech.

\begin{table*}
\renewcommand{\arraystretch}{1.2}
\small
  \centering
    \begin{tabular}{m{2.5cm}<{\centering}|m{1cm}<{\centering}m{1cm}<{\centering}|m{1cm}<{\centering}m{1cm}<{\centering}|m{1cm}<{\centering}m{1cm}<{\centering}|m{1cm}<{\centering}m{1cm}<{\centering}}
    \hline
          & \multicolumn{2}{c|}{Overall} & \multicolumn{2}{c|}{$A^{++}$} & \multicolumn{2}{c|}{$A^{+}$} & \multicolumn{2}{c}{$A^{o}$} \\
    \hline
    Model & \multicolumn{1}{c}{MSE$\;\downarrow$} & \multicolumn{1}{c|}{$\rho\uparrow$} & \multicolumn{1}{c}{MSE$\;\downarrow$} & \multicolumn{1}{c|}{$\rho\uparrow$} & \multicolumn{1}{c}{MSE$\;\downarrow$} & \multicolumn{1}{c|}{$\rho\uparrow$} & \multicolumn{1}{c}{MSE$\;\downarrow$} & \multicolumn{1}{c}{$\rho\uparrow$} \\
    \hline
    \multicolumn{9}{c}{\textit{Closed-Source Large Language Models (CS-LLMs)}} \\
    \hline
    GPT-4o & \textbf{0.1138}  & 0.6535  & \textbf{0.0514}  & \textbf{0.8098}  & \textbf{0.1268}  & \textbf{0.6298}  & \textbf{0.1928}  & 0.2332  \\
    GPT-4 & 0.1716  & \underline{0.6819}  & 0.1131  & 0.7175  & 0.2064  & 0.5323  & 0.2224  & \textbf{0.2478}  \\
    GPT-3.5 & 0.2163  & 0.5889  & 0.1878  & 0.6021  & 0.2430  & 0.4236  & 0.2309  & 0.1820  \\
    Claude-3.5 & \underline{0.1306}  & 0.6780  & \underline{0.0657}  & 0.7590  & \underline{0.1544}  & 0.5818  & \underline{0.2022}  & \underline{0.2379}  \\
    Gemini-1.5 & 0.2137  & 0.6305  & 0.1638  & 0.6970  & 0.2505  & 0.4517  & 0.2498  & 0.1877  \\
    \hline
    \textit{Avg. of CS-LLMs} & 0.1692  & 0.6466  & 0.1164  & 0.7171  & 0.1962  & 0.5238  & 0.2196  & 0.2177  \\
    \hline
    \multicolumn{9}{c}{\textit{Open-Source Large Language Models (OS-LLMs)}} \\
    \hline
    LLaMa3-70B & 0.1400  & \textbf{0.6990}  & 0.0753  & \underline{0.7634}  & 0.1680  & \underline{0.5856}  & 0.2072  & 0.2369  \\
    LLaMa3-8B & 0.1803  & 0.5912  & 0.1316  & 0.6533  & 0.2068  & 0.4572  & 0.2251  & 0.1667  \\
    Qwen2.5-72B & 0.1909  & 0.6588   & 0.1380  & 0.6817  & 0.2235  & 0.5001  & 0.2359  & 0.2119  \\
    Qwen2.5-7B & 0.1962  & 0.6024  & 0.1480  & 0.6638  & 0.2237  & 0.4756  & 0.2393  & 0.2056  \\
    Mixtral-8x22B & 0.1810  & 0.6287  & 0.1251  & 0.6858  & 0.2112  & 0.4944  & 0.2326  & 0.2107  \\
    Mixtral-8x7B & 0.1978  & 0.5921  & 0.1578  & 0.6267  & 0.2218  & 0.4709  & 0.2323  & 0.2132  \\
    \hline
    \textit{Avg. of OS-LLMs} & 0.1810  & 0.6287  & 0.1293  & 0.6791  & 0.2092  & 0.4973  & 0.2287  & 0.2075  \\
    \hline
    \end{tabular}%
\vspace{-0.025in}
\caption{Estimation of relationship between annotators and LLMs on MD-Agreement and its three subsets.
Results show Mean Squared Error (MSE) and Spearman's Rank Correlation Coefficient ($\rho$). }
\vspace{-0.075in}

  \label{main_2}%
\end{table*}%

\subsection{Evaluation of Relationship between Agreement Degree and LLM Confidence }\label{sec_main_2}

We analyze how well LLM confidence aligns with human annotation agreement, as a well-calibrated model should exhibit high confidence for clear cases and lower confidence for ambiguous cases. If LLMs assign high confidence to disagreement samples, this may indicate overconfidence, limiting their ability to reflect human-like uncertainty. To evaluate this, we apply the self-consistency method \cite{DBLP:conf/iclr/0002WSLCNCZ23, DBLP:conf/acl/Chen024}, which resamples model outputs under varying temperature settings to estimate confidence.

To measure confidence, we evaluate models under five temperature settings: 0, 0.25, 0.5, 0.75, and 1. Higher temperatures introduce more randomness in predictions, helping assess the model’s certainty across varying conditions. The final confidence score is computed by averaging the hard predictions across these temperature settings.

We use Mean Squared Error (MSE) to measure the alignment between LLM confidence and annotation agreement, where a smaller MSE indicates closer alignment \cite{uma-etal-2021-semeval, DBLP:conf/semeval/LeonardelliAABF23}. Additionally, we employ Spearman’s Rank Correlation Coefficient ($\rho$) to assess statistical correlation. The detailed metric definitions are provided in Appendix~\ref{formulas}. The results are presented in Table~\ref{main_2}.

%We then evaluate the relationship between LLM confidence and human annotation agreement. The self-consistency method \cite{DBLP:conf/acl/Chen024, DBLP:conf/iclr/0002WSLCNCZ23} is applied to measure the model's confidence by resampling its outputs with varying temperature coefficients. To ensure a robust evaluation, we perform experiments across five temperature settings: 0, 0.25, 0.5, 0.75, and 1. The final soft predictions are obtained by averaging the hard predictions from the outputs. Mean Squared Error (MSE) is used as the metric to evaluate the alignment degree between LLM confidence and annotation agreement, following \cite{uma-etal-2021-semeval, DBLP:conf/semeval/LeonardelliAABF23}, where a smaller MSE indicates closer alignment. We also use Spearman's Rank Correlation Coefficient ($\rho$) to assess the statistical correlation between the two.  The detailed introduction of metrics are provided in Appendix \ref{formulas}. The results are presented in Table \ref{main_2}. 

% The results of the alignment estimation between the degree of annotation agreement and the LLM confidence are presented in Table \ref{main_2}. Based on the results, we can draw the conclusions:

\begin{figure}
\centering
\includegraphics[width=7.25cm]{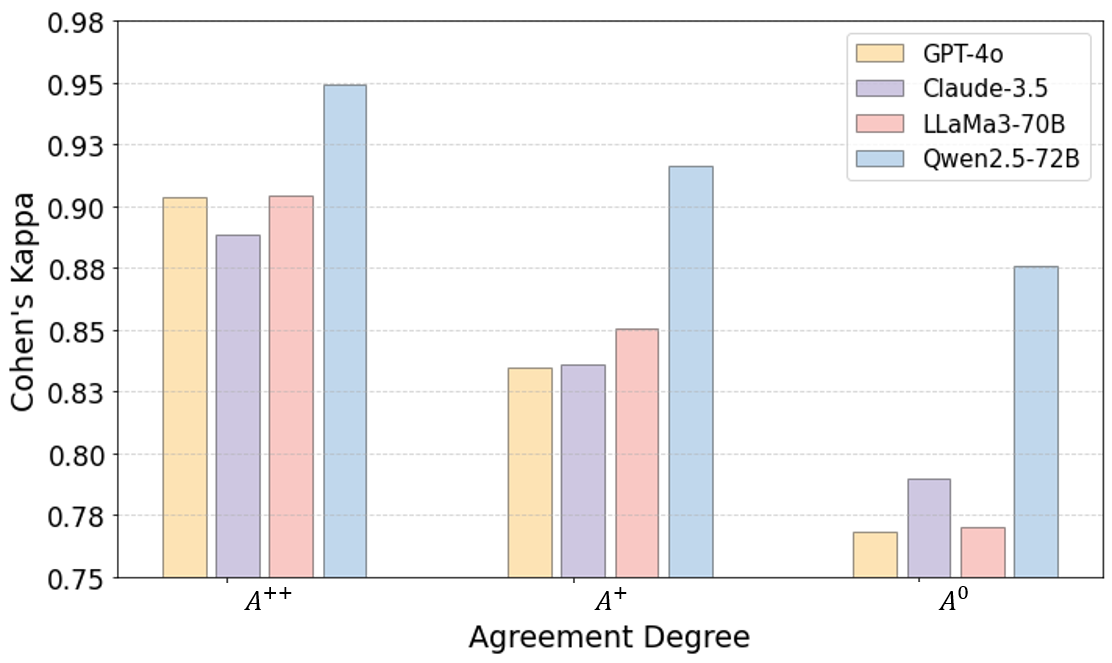}
\vspace{-0.075in}
\caption{Self-consistency of several LLMs across varying degrees of annotation agreement with Cohen's Kappa ($\kappa$) as the metric.}
\vspace{-0.125in}
\label{self_consistency}
\end{figure}

\paragraph{(1) As annotation agreement decreases, the alignment between model confidence and human agreement weakens.} As annotation agreement decreases, LLMs become less reliable in assessing their own uncertainty. GPT-4o, which performs best overall, has an MSE of 0.05 for $A^{++}$ samples but sees this error rise to 0.2 for $A^0$ samples. Additionally, Spearman’s correlation ($\rho$) between confidence and agreement weakens from above 0.7 for unanimous samples to below 0.3 for disagreement cases. This suggests that LLMs do not effectively recognize uncertainty in ambiguous cases. In real-world moderation, this could lead to overconfident misclassifications, where the model assigns a high confidence score to an incorrect label, making it harder to detect errors and apply human oversight.

\

%Based on the results, we observe that as annotation agreement decreases, the alignment degree between it and the LLM confidence shows a noticeable decline across multiple models. For example, GPT-4o, which demonstrates superior overall performance, has an MSE of about 0.05 for $A^{++}$ samples with unanimous agreement.  However, when detecting $A^{0}$ samples with low agreement, the MSE increases to 0.2. The statistical correlation between the model's confidence and annotation agreement also weakens accordingly.  For $A^{++}$ samples, the correlation is generally high ($\rho > 0.7$) for closed-source models, while for $A^{+}$ and $A^{0}$ samples with disagreements, the correlation is medium $(0.3 < \rho \leq 0.7$) and low ($\rho < 0.3$), respectively.
% These findings further suggest that existing LLMs face challenges in detecting samples with annotation disagreements and struggle to capture their subtle nature. These findings suggest that existing LLMs have not fully aligned with human standards for identifying offensive language, particularly for samples with annotation disagreement, where they struggle to capture the ambiguous nature.

\begin{figure}
\centering
\includegraphics[width=7.5cm]{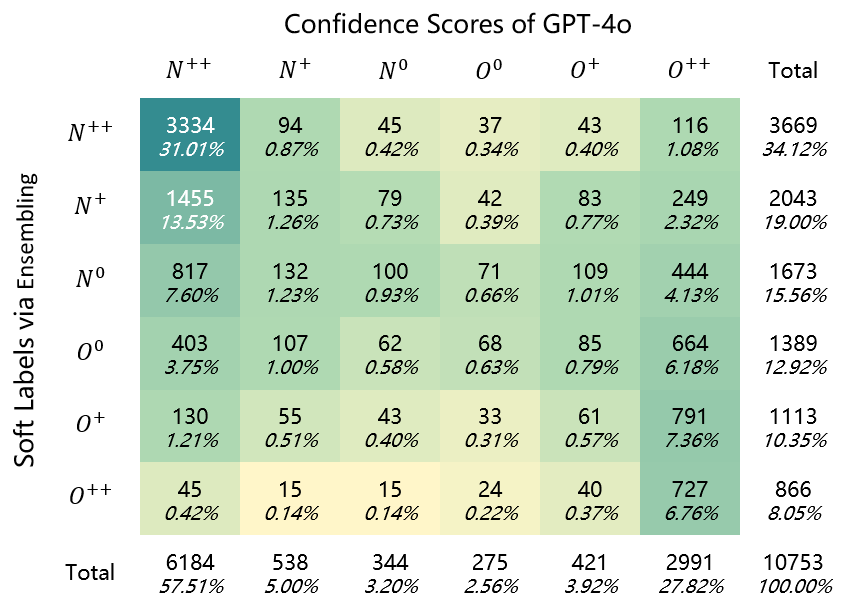}
\vspace{-0.075in}
\caption{Confusion matrix (raw counts and percentage) between confidence scores of GPT-4o (x-axis) and soft labels (y-axis).
}
\vspace{-0.175in}
\label{confusion}
\end{figure}

\vspace{-0.25in}
\paragraph{(2) LLMs demonstrate high self-consistency but may be overconfident in disagreement cases.} 
We assess self-consistency using Cohen’s Kappa ($\kappa$), measuring how stable LLM outputs remain across multiple sampling attempts. 
As shown in Figure~\ref{self_consistency}, self-consistency decreases for lower agreement samples but remains above 0.75 even for $A^0$ cases, indicating strong internal agreement (see Table A2). 
While high self-consistency is expected for unambiguous cases, it becomes problematic in ambiguous cases where human annotators disagree. 
This suggests that LLMs fail to capture the inherent uncertainty, exhibiting overconfidence by treating ambiguous inputs as clear-cut.
Such rigidity hinders their ability to adapt to nuanced linguistic and contextual variations.

\paragraph{(3) Even high-performing models exhibit overconfidence, limiting their ability to reflect human-like uncertainty.} 
We construct a confusion matrix of GPT-4o to visually analyze the relationship between the model’s confidence score and the soft labels of samples, as shown in Figure~\ref{confusion}.
The result reveals that even GPT-4o, the best-performing model, assigns high confidence to its predictions regardless of annotation agreement, indicating a lack of adaptability to disagreement cases.
This overconfidence highlights a critical flaw in LLM-based moderation: their inability to reflect the diversity of human judgment. Overconfident models are more likely to make systematic errors in handling subjective content, leading to unreliable moderation outcomes. 
Instead of relying on LLMs as sole decision-makers, future research should explore ensemble methods, uncertainty-aware training, or human-AI collaboration to mitigate biases and improve disagreement resolution.

We further analyze consistency across different models in Appendix B.2, and reveal low agreement among LLMs on disagreement samples.
This highlights the potential of ensemble models to handle these nuanced cases.

\begin{table*}
\small
\renewcommand{\arraystretch}{1.1}
  \centering
    \begin{tabular}{m{3cm}|m{0.925cm}<{\centering}m{1cm}<{\centering}|m{0.925cm}<{\centering}m{1cm}<{\centering}|m{1cm}<{\centering}m{1cm}<{\centering}|m{0.925cm}<{\centering}m{1cm}<{\centering}}
    \hline
          & \multicolumn{2}{c|}{Test Set} & \multicolumn{2}{c|}{$A^{++}$} & \multicolumn{2}{c|}{$A^{+}$} & \multicolumn{2}{c}{$A^{o}$} \\
    \hline
    \multicolumn{1}{c|}{Model} & Acc.$\;\uparrow$   & MSE$\;\downarrow$   & Acc.$\;\uparrow$   & MSE$\;\downarrow$   & Acc.$\;\uparrow$   & MSE$\;\downarrow$   & Acc.$\;\uparrow$   & MSE$\;\downarrow$ \\
    \hline
    \multicolumn{1}{c|}{GPT-4o (\textit{zero-shot})} & 80.11  & 0.1133  & 93.11  & 0.0560  & 79.76  & 0.1242  & 60.86  & \textbf{0.1923}  \\
    \hdashline[4pt/5pt]
    $\qquad \qquad w/ \; A^{++}$ & 80.93  & 0.1260  & 93.65  & 0.0647  & 81.41  & 0.1332  & 61.21  & 0.2107  \\
    $\qquad \qquad w/ \; A^{+}$ & 81.26  & 0.1144  & 93.65  & 0.0515  & 81.96  & 0.1157  & 61.80  & 0.2080  \\
    $\qquad \qquad w/ \; A^{0}$ & 81.22  & 0.1138  & 93.73  & 0.0558  & 82.18  & 0.1171  & 61.33  & \underline{0.1979}  \\
    \hdashline[4pt/5pt]
    $\qquad \qquad w/ \; A^{++/+}$ & 83.74  & \underline{0.1054}  & 95.28  & \textbf{0.0361}  & \underline{86.03}  & 0.1120  & \textbf{63.90}  & 0.2032  \\
    $\qquad \qquad w/ \; A^{++/0}$ & \textbf{83.87}  & 0.1063  & \textbf{95.74}  & 0.0416  & \textbf{86.14}  & \underline{0.1079}  & \underline{63.55}  & 0.2022  \\
    $\qquad \qquad w/ \; A^{+/0}$ & 82.07  & 0.1171  & 94.66  & 0.0500  & 82.18  & 0.1290  & 62.97  & 0.2059  \\
   $\qquad \qquad w/ \; A^{++/+/0}$ & \underline{83.51}  & \textbf{0.1045}  & \underline{95.51}  & \underline{0.0367}  & 85.48  & \textbf{0.1078}  & 63.32  & 0.2035  \\
    \hline
    \end{tabular}%
    \vspace{-0.025in}
  \caption{Performance of GPT-4o on the test set of MD-Agreement in few-shot learning: Accuracy (Acc.) for binary classification and MSE for evaluating alignment degree between annotation agreement and model's confidence. 
  % The table is divided into three sections based on prompt settings:
  The first row shows GPT-4o's performance in the zero-shot scenario, while the second and third sections evaluate the model with prompts containing a single level and combinations of agreement, respectively.}
 \vspace{-0.05in}
  \label{few_shot}%
\end{table*}%

%\section{How Do Disagreement Samples Affect LLMs During the Learning Phase?}
\section{RQ2: Impact of Disagreement Samples on LLM Learning}

In this section, we examine how samples with varying annotation agreements influence LLM performance during the learning phase. We focus on two key learning paradigms: few-shot learning and instruction fine-tuning. Specifically, we explore the impact of both single-category agreement samples and different agreement-level combinations on model performance.

\begin{table*}
\small
\renewcommand{\arraystretch}{1.1}
  \centering
    \begin{tabular}{m{3.5cm}|m{0.925cm}<{\centering}m{1cm}<{\centering}|m{0.925cm}<{\centering}m{1cm}<{\centering}|m{1cm}<{\centering}m{1cm}<{\centering}|m{0.925cm}<{\centering}m{1cm}<{\centering}}
    \hline
          & \multicolumn{2}{c|}{Test Set} & \multicolumn{2}{c|}{$A^{++}$} & \multicolumn{2}{c|}{$A^{+}$} & \multicolumn{2}{c}{$A^{o}$} \\
    \hline
    \multicolumn{1}{c|}{Model} & Acc.$\;\uparrow$   & MSE$\;\downarrow$   & Acc.$\;\uparrow$   & MSE$\;\downarrow$   & Acc.$\;\uparrow$   & MSE$\;\downarrow$   & Acc.$\;\uparrow$   & MSE$\;\downarrow$ \\
    \hline
    \multicolumn{1}{c|}{LLaMa3-8B (zero-shot)} & 70.92  & 0.1856  & 85.22  & 0.1350  & 66.45  & 0.2167  & 54.09  & 0.2288  \\
    \hdashline[4pt/5pt]
    $\qquad \qquad w/ \; A^{++}$ & 75.79  & 0.1671 & 89.01  & 0.1064  & 73.60  & 0.1919  & 58.18  & 0.2366  \\
    $\qquad \qquad w/ \; A^{+}$ & 77.04  & 0.1552 & 90.40  & 0.0898  & 74.81  & 0.1815  & 59.70  & 0.2291  \\
    $\qquad \qquad w/ \; A^{0}$ & 73.99  & 0.1348 & 86.53  & 0.1020  & 70.74  & 0.1537  & 56.31  & \textbf{0.1665}  \\
    \hdashline[4pt/5pt]
    $\qquad \qquad w/ \; A^{++/+}$ & 80.27  & 0.1340  & 93.34  & 0.0643  & 78.33  & 0.1582  & 60.86  & 0.2232  \\
    $\qquad \qquad w/ \; A^{++/0}$ & 79.29  & 0.1292 & 92.49  & 0.0641  & 78.11  & 0.1503  & 60.16  & 0.2138  \\
    $\qquad \qquad w/ \; A^{+/0}$ & \underline{82.53}  & \textbf{0.1075} & \underline{95.20}  & \underline{0.0404}  & \underline{83.94}  & \textbf{0.1160}  & \underline{61.80}  & \underline{0.1978}  \\
    $\qquad \qquad w/ \; A^{++/+/0}$ & \textbf{84.23}  & \underline{0.1106} & \textbf{95.98}  & \textbf{0.0379}  & \textbf{85.81}  & \underline{0.1186}  & \textbf{64.37}  & 0.2150  \\
    \hline
    \end{tabular}%
  % \vspace{-0.025in}
  \caption{Performance of LLaMa3-8B on the test set of MD-Agreement under instruction fine-tuning.
  % : Accuracy (Acc.) for binary classification and MSE for evaluating alignment degree between annotation agreement and model's confidence.
  }
  % \vspace{-0.05in}
  \label{fine-tuning}%
\end{table*}%

\subsection{Impact of Disagreement Samples on Few-Shot Learning}\label{sec_few-shot}

We evaluate the effect of disagreement samples on GPT-4o’s binary classification accuracy and its confidence alignment with human annotations during few-shot learning. %Understanding how disagreement samples contribute to model learning is crucial for optimizing prompt design in real-world applications.

%\paragraph{Few-Shot Learning Setup.}
%We construct prompts using positive and negative sample pairs randomly drawn from the MD-Agreement training set. Each prompt includes a fixed number of samples representing different levels of annotation agreement.
%Following \cite{DBLP:conf/emnlp/LeonardelliMAGT21}, we evaluate both single-category agreement samples and combinations of agreement levels.

 % number of samples representing different levels of annotation agreement.

%Specifically, prompts include different agreement configurations. The simplest setup, $w/\;A^{++}$, consists of only unanimous agreement ($A^{++}$) samples, each containing one offensive and one non-offensive example. The $w/\;A^{+/0}$ setup includes both medium agreement ($A^+$) and low agreement ($A^0$) samples, totaling four samples per prompt. We also examine mixed setups, such as $w/\;A^{++/0}$ and $w/\;A^{++/+}$, which combine unanimous agreement samples with one level of disagreement. Additionally, we assess a broader configuration, $w/\;A^{++/+/0}$, incorporating samples from all three agreement levels.
%\red{Junyu: Since $w/\;A^{+/0}$ also seems like a mixed setup, would it be more appropriate to present them together? In addition, the sample set used by combinations is composed of sample pairs of their respective single-category for reliable evaluation. Does this need to be emphasized? Here is another version inspired by yours:}

\paragraph{Few-Shot Learning Setup.} 
We evaluate both single-category agreement samples and combinations of agreement levels in few-shot learning, following \cite{DBLP:conf/emnlp/LeonardelliMAGT21}.
We first construct prompts using positive and negative sample pairs randomly drawn from the MD-Agreement training set, with each prompt including pairs corresponding to the respective agreement level.
For example, the simplest setup $w/\;A^{++}$ consists of only unanimous agreement ($A^{++}$) samples, containing one offensive and one non-offensive example.
Furthermore, we examine mixed setups with different agreement configurations, consisting of sample pairs from their respective single categories for reliable evaluation. 
For instance, $w/\;A^{++/0}$ and $w/\;A^{++/+}$ combine unanimous agreement samples with one level of disagreement, respectively. 
Additionally, we assess a broader configuration, $w/\;A^{++/+/0}$, which includes samples from all three agreement levels. The template details are provided in Appendix A.3.

We evaluate model performance on the MD-Agreement test set, analyzing both overall results and performance across different agreement levels. Table \ref{fine-tuning} summarizes the key findings.

\paragraph{(1) Few-shot learning improves classification accuracy but may increase overconfidence in ambiguous samples.}
Few-shot learning enhances classification accuracy, particularly in the medium agreement subset ($A^+$), where accuracy increases by an average of 3.87\%. However, for detection of low-agreement samples ($A^0$), few-shot learning increases the MSE, suggesting that models become overconfident and misaligned with ambiguous human annotations. This occurs because few-shot learning reinforces model consistency, making it less adaptable to subjective disagreements.

\paragraph{(2) Learning from disagreement samples improves model generalization.}
Using disagreement samples ($A^+$ and $A^0$) in few-shot learning leads to greater performance improvements across all evaluation metrics compared to using only unanimous agreement samples ($A^{++}$). Disagreement samples often capture borderline or ambiguous cases, which challenge the model to refine its decision boundaries. Learning from these samples enhances the model’s ability to differentiate nuanced offensive language from non-offensive content.

\paragraph{(3) Combining different agreement levels enhances performance, but excessive variation reduces accuracy.}
Incorporating both unanimous agreement samples and disagreement samples (e.g., $w/\;A^{++/0}$ or $w/\;A^{++/+}$) improves model performance compared to using only disagreement samples ($w/\;A^{+/0}$). However, including too many agreement categories ($w/\;A^{++/+/0}$) does not further enhance accuracy and may even decrease performance. The increased variation makes it harder for the model to establish clear decision boundaries, potentially leading to inconsistent classifications.
% \red{Junyu: For Qwen-72B (Table B1), the $w/\; A^{+/0}$ setup outperforms other combinations. This may be because the effects of learning from disagreement samples vary between different LLMs. This may affect some conclusions in (3) and (4). 
% }

%\paragraph{(4) A balanced mix of agreement samples optimizes few-shot learning.}
These results indicate that strategically balancing agreement levels is critical in few-shot learning. A well-chosen mix of clear and ambiguous cases helps the model generalize effectively, whereas excessive variation may introduce confusion and decrease performance.
%\red{Junyu: This seems to be a summary of the above findings, whether it is necessary to provide a separate number (4)}

To verify the robustness of our findings, we replicate the experiment using the open-source LLM Qwen2.5-72B. The results align closely with those of GPT-4o, suggesting that these insights generalize across different LLM architectures. Detailed results are provided in Appendix \ref{sec_few-shot_qwen}.

\subsection{Impact of Disagreement Samples on Instruction Fine-tuning}\label{sec_ft}
We analyze how instruction fine-tuning with different annotation agreement levels affects model performance, using LLaMa3-7B as the backbone.

\paragraph{Instruction Fine-tuning Setup.}
We fine-tune an equal number of instances from each agreement level in the MD-Agreement dataset. Specifically, we extract 1,800 samples each from $A^{++}$, $A^+$, and $A^0$, based on the least-represented $A^0$ category. The instruction template remains consistent with that used in the zero-shot setting (see Appendix \ref{template}). We also evaluate combinations of multiple agreement levels, using the same experimental markers as in Section \ref{sec_few-shot}. Table \ref{few_shot} presents the results, leading to the following conclusions:

\paragraph{(1) Medium-agreement ($A^+$) samples yield the best balance in fine-tuning.}  
Fine-tuning with high-agreement ($A^{++}$) samples improves classification accuracy, while low-agreement ($A^0$) samples enhance confidence alignment with human annotations, reducing MSE. However, exclusive reliance on $A^0$ samples may lead to catastrophic forgetting, where the model becomes overly attuned to ambiguous cases at the cost of general classification accuracy. $A^+$ samples offer the best trade-off, allowing the model to capture nuanced decision boundaries while maintaining robust performance.

\paragraph{(2) Combining multiple agreement levels further enhances performance.}  
Fine-tuning with all three agreement levels ($w/\;A^{++/+/0}$) achieves the best overall results, yielding performance comparable to GPT-4o in few-shot learning (see Table 5). Among two-category combinations, mixing disagreement samples ($w/\;A^{+/0}$) provides the most improvement, reinforcing the importance of disagreement-aware learning.

These results confirm that strategically selecting disagreement samples is essential for instruction fine-tuning. A well-balanced combination enhances both classification performance and confidence calibration, ensuring better alignment with human judgments. %\red{Junyu: Similarly, this seems to be a summary of the above findings, whether it is necessary to provide a separate number (7)}

We replicate the instruction fine-tuning experiment with Qwen2.5-7B using the same training and test data. The results closely align with those of LLaMa3-7B, confirming that these insights generalize across different model architectures (see Appendix \ref{sec_ft_qwen}).
Additionally, we conduct an in-depth analysis of the generalization ability of the trained models on out-of-distribution (OOD) data using disagreement samples, drawing on the bias-variance trade-off theory (see Appendix \ref{OOD}).

\section{Related Work}

\paragraph{Large Language Model.}
In recent years, large language models (LLMs) have rapidly emerged, showcasing extensive world knowledge and strong reasoning capabilities \cite{DBLP:conf/nips/KojimaGRMI22}. 
%  evaluates the ability of LLMs in non-deterministic question answering, aiming 
Many researchers have proposed diverse tasks to deeply analyze the relationship between the model's outputs and human judgments \cite{DBLP:journals/corr/abs-2408-01419, DBLP:journals/corr/abs-2404-05264}.
In addition, the confidence of LLMs in their outputs has also attracted attention from researchers, which is often used to assess the reliability and robustness of the generated content \cite{DBLP:journals/tacl/JiangADN21}.
Various methods for estimating confidence have been proposed \cite{DBLP:conf/iclr/0002WSLCNCZ23, DBLP:conf/emnlp/TianMZSRYFM23, DBLP:journals/tmlr/LinHE22}. 
In this study, we employ the most straightforward approach, \textit{self-consistency} \cite{DBLP:conf/iclr/0002WSLCNCZ23}, to estimate the model’s confidence.
% In this study, we reference related methods to assess the relationship between the degree of annotation agreement and the LLM confidence.

\paragraph{Offensive Language Detection.}
Researchers have developed various methods for detecting offensive language \cite{DBLP:conf/icwsm/FountaDCLBSVSK18, DBLP:conf/icwsm/DavidsonWMW17, DBLP:conf/aaai/MathewSYBG021}.
As research advances, many studies argue that treating offensive language detection as a binary classification is an idealized assumption \cite{basile-etal-2021-need, DBLP:conf/aiia/Basile20, plank-2022-problem}, as annotation disagreement is inherent in datasets for such subjective task \cite{DBLP:journals/tacl/PavlickK19, DBLP:journals/jair/UmaFHPPP21}. 
Using majority voting for annotation agreement leads to information loss \cite{DBLP:journals/tacl/DavaniDP22}, as these disagreements arise from the subtlety of the samples, not labeling errors \cite{DBLP:journals/frai/UmaAP22}. 
\citet{DBLP:conf/semeval/LeonardelliAABF23} emphasizes that detection models should recognize this disagreement, rather than just improving classification performance.

Recently, several studies have begun evaluating the potential of LLMs for detecting offensive language \cite{DBLP:conf/icwsm/0006AD24, DBLP:conf/emnlp/RoyH0S23}, and designing detection methods based on them \cite{DBLP:conf/emnlp/ParkKJPH24, DBLP:conf/emnlp/WenKSZLBH23}. 
Some studies \cite{DBLP:conf/ijcai/WangHACL23, DBLP:conf/www/HuangKA23a} leverage the generative capabilities of LLMs to provide explanations for offensive language, assisting human annotation. 
Furthermore, \citet{DBLP:journals/corr/abs-2410-07991, DBLP:conf/acl/ZhangHJL24} assess the sensitivity of LLMs to demographic information in the context of offensive language.
Though great efforts have been made, these studies lack focus on the phenomenon of offensive language with annotation disagreement.
In this paper, we aim to fill this research gap.

\section{Conclusion}
This study examines how LLMs handle annotation disagreement in offensive language detection, a critical challenge in real-world moderation. We evaluate multiple LLMs in a zero-shot setting and find that while they perform well on unanimously agreed-upon samples, their accuracy drops significantly for disagreement cases. Moreover, their overconfidence leads to rigid predictions, misaligning them with human annotations.

To address this, we investigate the impact of disagreement samples in few-shot learning and instruction fine-tuning. Our results show that incorporating these samples improves detection accuracy and human alignment, enabling LLMs to better capture the subjective nature of offensive language. We further find that balancing agreement levels in training data prevents overfitting to ambiguous cases, ensuring model robustness.

Key findings of this work include: (1) a systematic evaluation of LLMs on annotation disagreement, (2) insights into how disagreement samples improve learning, and (3) guidelines for leveraging disagreement-aware training strategies. These results emphasize the need for model calibration techniques to mitigate overconfidence and for training strategies that incorporate disagreement to improve generalization. 

Future research should explore dynamic fine-tuning approaches and confidence-aware moderation systems to bridge the gap between LLM decisions and human subjectivity.
We hope that our findings and insights will contribute to the further development of offensive language detection, LLM-as-a-judge, and human-AI collaboration.

\newpage
\section*{Limitations and Ethics Statement}

(1) Due to the scarcity of high-quality offensive language datasets with unaggregated labels, we only utilize the MD-Agreement dataset for experiments, which has been widely used in the field. 
Considering that relying on a single dataset may introduce bias or randomness, we mitigate this by conducting experiments with multiple closed-source and open-source LLMs to ensure the consistency and reliability of our findings, reducing the impact of bias. 
In future work, we plan to further explore the performance of LLMs in other subjective text analysis tasks, such as humor detection and misogyny detection, particularly in understanding samples with annotation disagreement.

(2) Due to usage restrictions, we are unable to evaluate the detection performance of several emerging LLMs, such as GPT-o3. We plan to further assess these more advanced models as soon as experimental conditions allow.
Additionally, due to space limitations, the potential of certain techniques for detecting offensive language with annotation disagreement, such as reinforcement learning methods, are not discussed. 
We plan to explore these methods in future work and investigate effective strategies for enabling LLMs to fully leverage disagreement samples, thereby enhancing their detection capabilities.
% The contribution of this paper primarily lies in constructing benchmarks and evaluating the effectiveness of existing disagreement learning methods. 
% , and learn the boundaries between offensive and non-offensive samples, 

(3) In evaluating the confidence of LLMs, we adopt a straightforward approach based on temperature resampling, which has been widely used in recent LLM literature, especially for evaluating uncertainty in black-box settings, where internal logits or token probabilities may not be accessible.
We fully acknowledge the diversity of uncertainty metrics and plan to explore logit-based calibration methods \cite{DBLP:conf/icml/GuoPSW17} (e.g., ECE, Brier score, and temperature scaling) and Bayesian approximations in future work to deepen our understanding of LLM uncertainty in handling annotation disagreements.

(4) The opinions and findings contained in the samples of this paper should not be interpreted as representing the views expressed or implied by the authors. 
Accessing the MD-Agreement dataset requires users to agree to the creators' usage agreements. 
The usage of these samples in this study fully complies with these agreements \footnote{The code of this work is available at \url{https://github.com/DUT-lujunyu/Disagreement}}.

% \section*{Ethics Statement}

% We hope this work can enhance researchers' attention to offensive language with annotation disagreements, as these samples are more challenging and subtle.

\section*{Acknowledgments}
This research is supported by the Natural Science Foundation of China (No. 62376051), the Liaoning Provincial Natural Science Foundation Joint Fund Program (2023-MSBA-003), the Fundamental Research Funds for the Central Universities (DUT24LAB123, DUT24MS003), and the Ministry of Education, Singapore, under its Academic Research Fund Tier 2 (Award ID: MOE-T2EP20222-0010). 
Any opinions, and conclusions or recommendations expressed in this material are those of the authors and do not reflect the views of the National Natural Science Foundation of China and the Ministry of Education, Singapore.

\bibliography{uncertainty, LLM_hate, LLM_base, basic_hate}

\newpage

\appendix

\renewcommand\thefigure{\Alph{section}\arabic{figure}}    
\renewcommand\thetable{\Alph{section}\arabic{table}} 

\section{Experimental Details}

\subsection{Details of Dataset}\label{data_details}

In this section, we provide a detailed introduction to the annotation quality control process of our used MD-Agreement dataset \cite{DBLP:conf/emnlp/LeonardelliMAGT21}.
The researchers implemented a two-stage annotation process: First, three linguists annotated a subset of the samples, and those with unanimous agreement were used as the gold standard for the annotation process. Following this, trained annotators from Amazon Mechanical Turk were employed to annotate the complete samples based on the established gold standard.
After the task was completed, annotations from workers who did not achieve at least 70\% accuracy were discarded. Additionally, it was ensured that each sample in the final dataset received five annotations. These measures help ensure the accuracy of the annotations.
\citet{DBLP:conf/eacl/SandriLTJ23} further manually reviewed a random selection of 2,570 samples with annotation disagreement from the MD-Agreement dataset. The results showed that only 12 samples contained annotation errors, accounting for less than 0.5\%, demonstrating the high quality and reliability of the dataset.

% \subsection{Examples with Varying Agreement Levels}\label{Examples}

% To facilitate the understanding of samples with different degrees of annotation disagreement, we present several such samples for reference, as shown in Table \ref{examples}.

\subsection{Description of Metrics}\label{formulas}

This section introduces the metrics used to assess the relationship between LLM confidence and the degree of human annotation agreement.

\textbf{Mean Squared Error (MSE)}:
The MSE is a widely used evaluation metric in regression tasks, measuring the difference between predicted and actual values.
In this study, we adopt MSE for alignment estimation, as described by  \citet{DBLP:conf/semeval/LeonardelliAABF23}, where a smaller MSE indicates closer alignment between LLM confidence and agreement degree.
% We first convert the discrete 0-1 sequences of annotations and LLM's predictions under different sampling of samples into continuous probability values through averaging to obtain the soft labels and soft predictions, as follows:
We first obtain soft labels $y$ and soft predictions $\hat{y}$ of samples by averaging their discrete 0-1 annotation sequences $Y$ and the LLM outputs $\hat{Y}$ across different samplings, as follows:
\begin{equation}
y_i = \frac{1}{n} \sum_{i=1}^{n} Y_i, \quad \hat{y_i} = \frac{1}{n}. \sum_{i=1}^{n} \hat{Y_i},  
\end{equation}
where $n$ is the number of observations, set to $n = 5$ in this paper, representing the number of annotators and LLM outputs. 
Then, the MSE is calculated as:
\begin{equation}
MSE = \frac{1}{m} \sum_{i=1}^{m} (y_i - \hat{y}_i)^2,
\end{equation}
where $m$ is the total number of samples.

\setcounter{table}{0}
\begin{table}
\small
\renewcommand{\arraystretch}{1.2}
  \centering
    \begin{tabular}{cc}
    \toprule
    Range of Coefficient (($\rho$) & Correlation Degree \\
    \midrule
    $(0.7, 1.0]$ & High Correlation \\
    $(0.3, 0.7]$ & Medium Correlation \\
    $(0.0, 0.3]$ & Low Correlation \\
    $0.0$   & No Correlation \\
    $[-1.0, 0.0)$ & Negative Correlation \\
    \bottomrule
    \end{tabular}%
  \caption{Correlation degree corresponding to different coefficient values ($\rho$).}
    \label{coefficient}%
\end{table}%

\begin{table}
\small
\renewcommand{\arraystretch}{1.2}
  \centering
    \begin{tabular}{cc}
    \toprule
    Range of Kappa ($\kappa$) & Agreement Degree \\
    \midrule
    $[0.8, 1.0]$ & High Agreement \\
    $[0.6, 0.8)$ & Good Agreement \\
    $[0.4, 0.6)$ & Moderate Agreement \\
    $(0.0, 0.4)$ & Poor Agreement \\
    $0.0$   & No Agreement \\
    $[-1.0, 0.0)$ & Negative Correlation \\
    \bottomrule
    \end{tabular}%
  \caption{Agreement degree corresponding to different kappa values ($\kappa$).}
  \label{kappa}%
\end{table}%

\textbf{Spearman’s Rank Correlation Coefficient ($\rho$):}
The Spearman’s Rank Correlation Coefficient is a non-parametric test that quantifies the degree of monotonic relationship between two variables. 
Unlike Pearson correlation, which assumes normally distributed variables, Spearman's correlation does not require this assumption and can be applied to discrete data. 
This makes it an ideal choice for assessing the statistical correlation between annotation agreement and LLM confidence, which is computed as follows:

\begin{equation}
\rho = 1 - \frac{6 \sum_{i=1}^{n} d_i^2}{n(n^2 - 1)},
\end{equation}
where $di$ is the difference between the ranks of corresponding values of soft labels $y$ and predictions $\hat{y}$.
The correlation degrees corresponding to different $\rho$ are shown in Table \ref{coefficient}.

\textbf{Cohen's Kappa ($\kappa$)}: 
Cohen's Kappa is widely used to evaluate the consistency between annotators, especially in classification tasks. 
Compared to simple percentage agreement, which directly calculates the proportion of agreement between two evaluators, Cohen's Kappa provides a more precise measure as it accounts for the influence of random agreement. 
% In this paper, we treat the LLM as an alternative annotator for offensive language, making Cohen's Kappa a suitable choice to estimate the consistency between annotator decisions $Y$ and LLM outputs $\hat{Y}$, which are two 0-1 sequences. 
In this paper, we use Cohen's Kappa to estimate the consistency of LLM outputs $\hat{Y}$. 
We first compute the kappa value between the $i$-th and the $j$-th LLM output of samples, i.e., $\hat{Y_i}$ and $\hat{Y_j}$, using the following formula:

\begin{equation}
\kappa_{i, j} = \frac{P_{o_{i, j}} - P_{e_{i, j}}}{1 - P_{e_{i, j}}},
\end{equation}
where $P_{o_{i, j}}$ represents the observed agreement, which is the proportion of agreement between $Y_i$ and $Y_j$, and $P_{e_{i, j}}$ refers to the expected agreement, calculated based on the probability of selecting categories, namely 0 or 1.
We then calculate the average value of the kappas as follows, which is used as the metric:
\begin{equation}
\kappa = \frac{1}{C(n,2)} \sum_{1\leq i<j \leq 5} \kappa_{i, j}
\end{equation}
The agreement degrees corresponding to different $\kappa$ are shown in Table \ref{kappa}.

\begin{table}[t]
\small
\renewcommand{\arraystretch}{1.2}
  \centering
    \begin{tabular}{p{22.5em}}
    \toprule
    Offensive language refers to words or expressions intended to insult, harm, or belittle individuals or groups. This includes:\newline{}
    \quad $\cdot$ Insulting, derogatory, or combative language\newline{}
    \quad $\cdot$ Discriminatory language based on race, religion, gender, or other personal attributes\newline{}
    \quad $\cdot$ Profanity, violence, or vulgar expressions\newline{}
    \quad $\cdot$ Sarcastic or harmful humor\newline{}
    Based on the above definition, determine if the following sentence is offensive language. Return only “\textit{yes}” or “\textit{no}”.\newline{}\newline{}
    <\textit{If in the Few-shot Learning}>\newline{}
    Example\_1: \newline{}
    Input: [\texttt{text of example\_1}]\newline{}
    Output: [\texttt{label\_1}]\newline{}
    Example\_2: \newline{}
    Input: [\texttt{text of example\_2}]\newline{}
    Output: [\texttt{label\_2}]\newline{}
    <\textit{Other Examples}>\newline{}\newline{}
    Here is the sample to be detected:\newline{}
    Input: [\texttt{sample to be detected}]\newline{}
    Output: [\texttt{prediction}]\\
    \bottomrule
    \end{tabular}%
    % \vspace{-0.025in}
  \caption{Prompt template of the LLM, consisting primarily of three parts: task definition, examples (only for the few-shot scenario), and the sample to be detected.}
  \label{template_table}%
\end{table}%

\subsection{Design of Prompt Template}\label{template}

To enhance the reproducibility of our study, we avoided conducting complex prompt engineering. 
Instead, we directly referenced \cite{DBLP:conf/emnlp/RoyH0S23} to design a straightforward prompt template, as shown in Table \ref{template_table}.
The template includes three parts: first, the definition of offensive language, which aligns with that used in the MD-Agreement dataset \cite{DBLP:conf/emnlp/LeonardelliMAGT21} to ensure the accuracy of the evaluation; 
second, examples of varying degrees of disagreement in a few-shot scenario; 
and finally, the sample to be detected.

\begin{table}
\small
\renewcommand{\arraystretch}{1.15}
  \centering
    \begin{tabular}{m{2cm}<{\centering}m{4.25cm}<{\centering}}
    \toprule
    Model & Version \\
    \midrule
    GPT-o1 & \texttt{o1-preview-2024-09-12} \\
    GPT-4o & \texttt{gpt-4o-2024-08-06} \\
    GPT-4 & \texttt{gpt-4-turbo-2024-04-09} \\
    GPT-3.5 & \texttt{gpt-3.5-turbo-0125} \\
    Claude-3.5 & \texttt{claude-3-5-sonnet-20240620} \\
    % Claude-3 & \texttt{claude-3-sonnet} \\
    Gemini-1.5 & \texttt{gemini-1.5-pro}  \\
    \hdashline[4pt/5pt]
    LLaMa3-70B & \texttt{Meta-Llama-3-70B-Instruct} \\ 
    LLaMa3-8B & \texttt{Meta-Llama-3-8B-Instruct} \\ 
    Qwen2.5-72B & \texttt{Qwen2.5-72B-Instruct} \\ 
    Qwen2.5-7B & \texttt{Qwen2.5-7B-Instruct} \\ 
    Mixtral-8x22B & \texttt{Mixtral-8x22B-Instruct-v0.1} \\ 
    Mixtral-8x7B & \texttt{Mixtral-8x7B-Instruct-v0.1} \\ 
    \bottomrule
    \end{tabular}%
    \vspace{-0.05in}
  \caption{Specific versions of used LLMs.}
\vspace{-0.1in}
  \label{LLM_version}%
\end{table}%

\subsection{Other Experimental Settings}\label{version}

We access closed-source LLMs via their official APIs and deploy open-source LLMs with parameters downloaded from Hugging Face. 
To ensure a fair comparison, we use model versions released around the same time, as detailed in Table \ref{LLM_version}.
Since GPT-o1 only has a default temperature of 1 and does not allow adjustments, we present its binary performance in this setting.
Except for the temperature coefficient, other hyperparameters, such as top-p and top-k, are set to their default values for each model. 
For instruction fine-tuning, we adopt the efficient Qlora fine-tuning method. 
The learning rate is set to 2e-4, with a per-device batch size of 36.
We train the model for 15 epochs using the AdamW optimizer, applying an early stopping mechanism. 
We reserve the parameters of best-performing models based on the development set and evaluate their performance on the test set. 
The models are trained on two NVIDIA H100 80GB GPUs.
All the few-shot learning and instruction fine-tuning experiments are repeated five times with different random seeds to minimize error, and the average results are reported.

\subsection{Handling of Refusal Behavior}
Handling offensive language can trigger the refusal behavior of LLMs, as they are designed with ethical and safety considerations \cite{DBLP:conf/icwsm/0006AD24}. 
Nevertheless, in our experiments, refusal occurred only in the zero-shot evaluation setting, where Claude-3.5, with a temperature coefficient set to 1, failed to generate responses for 23 samples.
When the experiment was repeated with the same settings, the model successfully provided predictions for these samples. 
This phenomenon also highlights the model's sensitivity to offensive language.
% and underscores the importance of incorporating multi-temperature coefficient sampling.

\section{Supplementary Experiments}

\setcounter{figure}{0}
\begin{figure*}
\centering
\includegraphics[width=16cm]{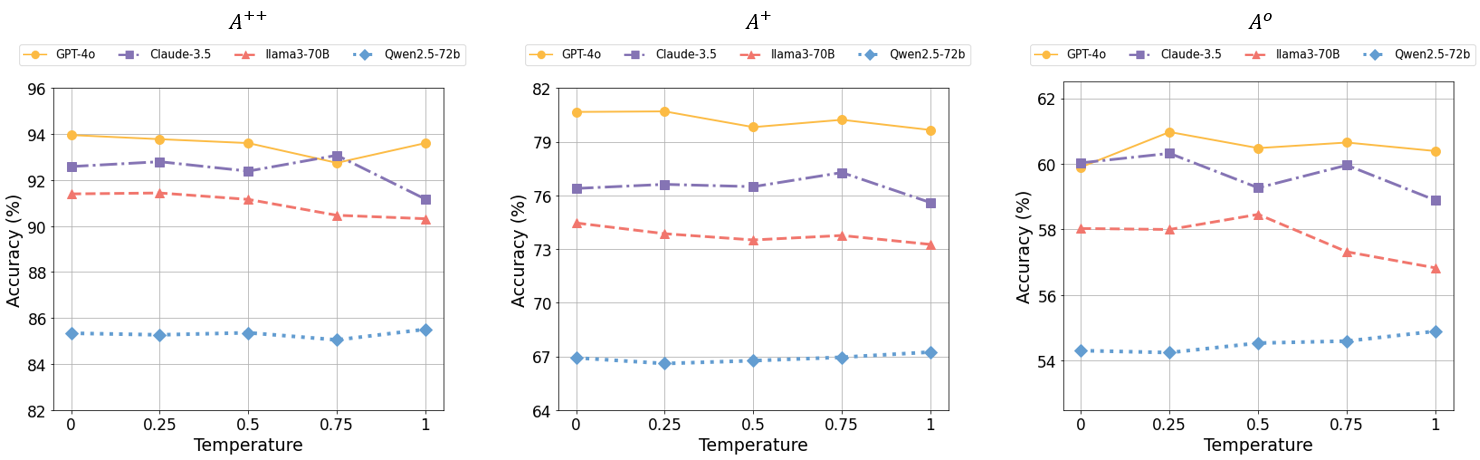}
\vspace{-0.25in}
\caption{Accuracy of LLMs on detecting offensive language with different degrees of annotation agreement under different temperature sampling settings.}
\vspace{-0.075in}
\label{temperature}
\end{figure*}

\begin{figure*}[t]
\centering
\includegraphics[width=16cm]{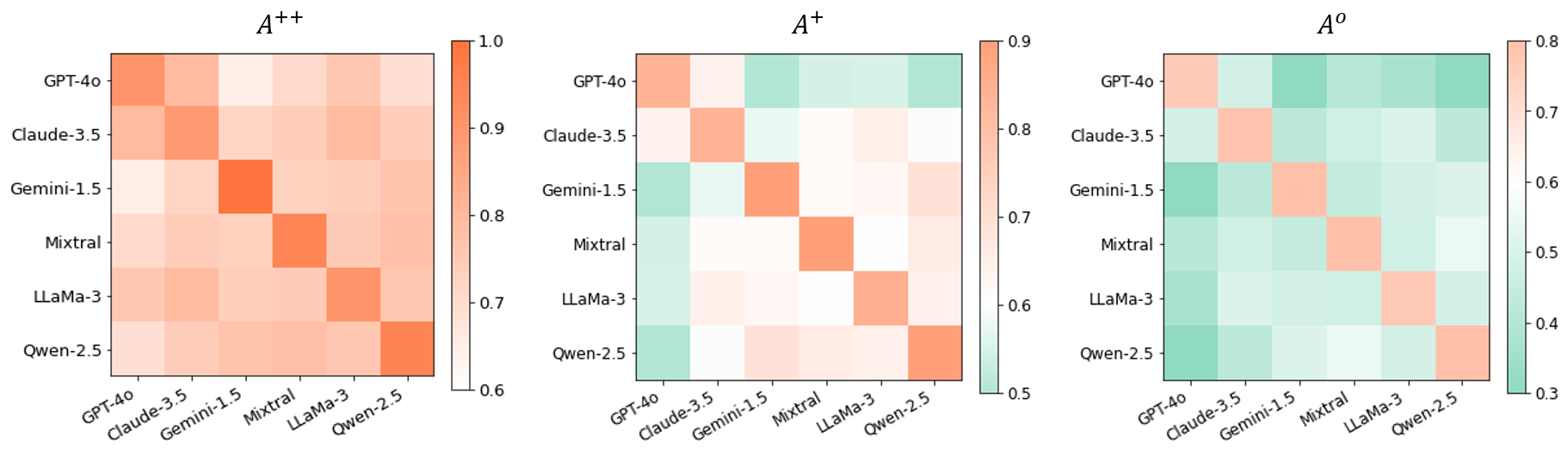}
\vspace{-0.275in}
\caption{Consistency of outputs from different LLMs across varying degrees of annotation agreement with Cohen's Kappa as the metric. The color scale represents different Kappa values.}
\vspace{-0.025in}
\label{dis_kappa}
\end{figure*}

\subsection{Impact of Temperature Sampling on Detection Performance of LLMs}

In this section, we analyze the impact of temperature sampling on the accuracy of detecting offensive language by LLMs. 
We select four representative models for comparison: the closed-source models GPT-4 and Claude-3.5, as well as the open-source models LLaMa3-70B and Qwen2.5-72B. 
The experimental results are shown in Figure \ref{temperature}.
Based on these results, we conclude that after adjusting the temperature coefficient, the detection accuracy of each LLM remains generally stable, although some fluctuations are observed, with varying degrees of sensitivity to the temperature coefficient across different models. 
As the temperature increases, the accuracy of most models shows a declining trend, with the sole exception being Qwen2.5-72B, which exhibits an increase in accuracy. 
This may be due to differences in the models' training mechanisms. 
Nevertheless, the performance ranking between the models remains stable, indicating that changes in the temperature coefficient do not notably affect the performance differences among the models.

\subsection{Consistency Analysis Across Different LLMs}

Building upon Section \ref{sec_main_2}, we further explore the consistency of hard predictions across different LLMs when processing samples with varying degrees of annotation agreement.
We select six representative models, including the close-source models GPT-4o, Claude-3.5, and Gemini-1.5, as well as the open-source models Mixtral-8x22B, LLaMa3-70B, and Qwen2.5-72B.
Cohen's Kappa is used as the metric.
The results are presented in Figure \ref{dis_kappa}.
Based on the results, we can observe that:

As annotation agreement decreases, cross-model consistency in detecting offensive language declines more significantly compared to each model’s self-consistency.
For unanimous agreement samples ($A^{++}$), cross-model consistency generally exhibits good agreement, with $\kappa > 0.6$. 
However, for low agreement samples $A^0$, consistency drops explicitly, with many models showing poor agreement ($\kappa < 0.4$), despite many of these models exhibiting similar overall performance in terms of both binary classification accuracy and alignment with human annotations (see Table \ref{main} and \ref{main_2}). 
Notably, the lowest prediction consistency Kappa is only 0.28 between GPT-4o and Gemini 1.5.
A potential reason for this phenomenon is that different models are trained on diverse datasets and undergo distinct value alignment processes, resulting in varying sensitivity to contextual features.
In future work, we will explore the relationship between cross-model consistency and human annotation agreement in offensive language detection. 
Additionally, we aim to investigate the potential of leveraging multiple LLMs for collaborative annotation of offensive language.

\setcounter{table}{0}
\begin{table*}
\small
\renewcommand{\arraystretch}{1.2}
  \centering
    \begin{tabular}{m{3.75cm}|m{0.925cm}<{\centering}m{1cm}<{\centering}|m{0.925cm}<{\centering}m{1cm}<{\centering}|m{1cm}<{\centering}m{1cm}<{\centering}|m{0.925cm}<{\centering}m{1cm}<{\centering}}
    \hline
          & \multicolumn{2}{c|}{Overall} & \multicolumn{2}{c|}{$A^{++}$} & \multicolumn{2}{c|}{$A^{+}$} & \multicolumn{2}{c}{$A^{o}$} \\
    \hline
    \multicolumn{1}{c|}{Model} & Acc.$\;\uparrow$   & MSE$\;\downarrow$   & Acc.$\;\uparrow$   & MSE$\;\downarrow$   & Acc.$\;\uparrow$   & MSE$\;\downarrow$   & Acc.$\;\uparrow$   & MSE$\;\downarrow$ \\
    \hline
    \multicolumn{1}{c|}{Qwen2.5-72B (\textit{zero-shot})} & 72.08 & 0.1962  & 84.74 & 0.1480  & 68.41 & 0.2237  & 57.12 & 0.2393  \\
    \hdashline[4pt/5pt]
    $\qquad \qquad w/ \; A^{++}$ & 77.92  & 0.1321  & 90.94  & 0.0809  & 75.91  & 0.1484  & 60.40  & 0.1920  \\
    $\qquad \qquad w/ \; A^{+}$ & 79.10  & 0.1275  & 91.95  & 0.0702  & 78.66  & 0.1414  & 60.16  & 0.1993  \\
    $\qquad \qquad w/ \; A^{0}$ & \underline{82.56}  & \underline{0.1054}  & \underline{94.12}  & \underline{0.0514}  & 83.50  & \textbf{0.1088}  & \textbf{64.14}  & \textbf{0.1832}  \\
    \hdashline[4pt/5pt]
    $\qquad \qquad w/ \; A^{++/+}$ & 81.42  & 0.1127  & 93.19  & 0.0561  & 82.51  & 0.1199  & 62.50  & \underline{0.1905}  \\
    $\qquad \qquad w/ \; A^{++/0}$ & 82.43  & 0.1099  & 93.58  & 0.0530  & \underline{84.38}  & 0.1108  & \underline{63.55}  & 0.1950  \\
    $\qquad \qquad w/ \; A^{+/0}$ & \textbf{82.96}  & \textbf{0.1044}  & \textbf{94.97}  & \textbf{0.0427}  & \textbf{84.71}  & \underline{0.1090}  & 62.97  & 0.1927  \\
    $\qquad \qquad w/ \; A^{++/+/0}$ & 82.04  & 0.1101  & 93.42  & 0.0544  & 84.05  & 0.1095  & 62.73  & 0.1949  \\
    \hline
    \end{tabular}%
    % \vspace{-0.025in}
  \caption{Performance of Qwen2.5-72B on the test set of MD-Agreement in few-shot learning.}
    % \vspace{-0.025in}
  \label{few-shot_qwen}%
\end{table*}%

\begin{table*}
\small
\renewcommand{\arraystretch}{1.2}
  \centering
    \begin{tabular}{m{3.75cm}|m{0.925cm}<{\centering}m{1cm}<{\centering}|m{0.925cm}<{\centering}m{1cm}<{\centering}|m{1cm}<{\centering}m{1cm}<{\centering}|m{0.925cm}<{\centering}m{1cm}<{\centering}}
    \hline
          & \multicolumn{2}{c|}{Overall} & \multicolumn{2}{c|}{$A^{++}$} & \multicolumn{2}{c|}{$A^{+}$} & \multicolumn{2}{c}{$A^{o}$} \\
    \hline
    \multicolumn{1}{c|}{Model} & Acc.$\;\uparrow$   & MSE$\;\downarrow$   & Acc.$\;\uparrow$   & MSE$\;\downarrow$   & Acc.$\;\uparrow$   & MSE$\;\downarrow$   & Acc.$\;\uparrow$   & MSE$\;\downarrow$ \\
    \hline
    \multicolumn{1}{c|}{Qwen2.5-7B (\textit{zero-shot})} & 69.77  & 0.1998  & 83.44  & 0.1542  & 66.12  & 0.2289  & 53.04  & 0.2379  \\
    \hdashline[4pt/5pt]
    $\qquad \qquad w/ \; A^{++}$ & 80.18 & 0.1407 & 92.96 & 0.0703 & 79.76 & 0.1572 & 61.21 & 0.2309 \\
    $\qquad \qquad w/ \; A^{+}$ & 80.41 & 0.1347 & 93.11 & 0.0649 & 80.64 & 0.1497 & 62.73 & 0.2211 \\
    $\qquad \qquad w/ \; A^{0}$ & 80.08 & 0.1395 & 92.57 & 0.0704 & 80.09 & 0.1533 & 60.86 & 0.2238 \\
    \hdashline[4pt/5pt]
    $\qquad \qquad w/ \; A^{++/+}$ & \underline{82.30} & 0.1261 & \underline{95.36} & \underline{0.0457} & 82.95 & 0.1406 & 62.38 & 0.2311 \\
    $\qquad \qquad w/ \; A^{++/0}$ & 82.17 & 0.1192 & 94.50 & 0.0486 & \underline{84.27} & \underline{0.1243} & \underline{63.90} & \underline{0.2082} \\
    $\qquad \qquad w/ \; A^{+/0}$ & 81.88 & \underline{0.1185} & 94.43 & 0.0522 & 83.39 & 0.1283 & \underline{63.90} & \textbf{0.2060} \\
    $\qquad \qquad w/ \; A^{++/+/0}$ & \textbf{83.91} & \textbf{0.1149} & \textbf{95.82} & \textbf{0.0395} & \textbf{85.59} & \textbf{0.1209} & \textbf{65.42} & 0.2181 \\
    \hline
    \end{tabular}%
    % \vspace{-0.025in}
  \caption{Performance of Qwen2.5-7B on the test set of MD-Agreement under instruction fine-tuning.}
    % \vspace{-0.025in}
  \label{ft_qwen}%
\end{table*}%

\subsection{Few-shot Learning with Qwen2.5-72B}\label{sec_few-shot_qwen}

We replicate the few-shot learning experiment from Section \ref{sec_few-shot} using the open-source LLM Qwen2.5-72B, employing the same sample pairs in the prompts.
The results are shown in Table \ref{few-shot_qwen}.
Based on the results, we observe the following:

In the few-shot learning with samples of varying annotation agreement degrees, the results of Qwen2.5-72B align closely with the trends of GPT-4o (see Table \ref{few_shot}). 
Whether introducing samples with a single annotation agreement degree or combinations of different agreement categories, the detection performance of the model shows notable improvement compared to the zero-shot scenario. 
Additionally, the benefit to model performance varies explicitly depending on the annotation agreement degree and the combinations used as prompts.

Furthermore, compared to GPT-4o, Qwen2.5-72B demonstrates two distinct differences:
(1) On the subset of low-agreement samples, the introduction of few-shot learning also results in a noticeable improvement in the alignment between LLM confidence and annotation agreement. 
This is primarily because Qwen2.5-72B performs less effectively than GPT-4o in detecting offensive language under zero-shot scenarios, making it more susceptible to performance enhancements through few-shot learning.
(2) For Qwen2.5-72B, the combination of disagreement samples, i.e., $w/\; A^{+/0}$, achieves the best performance across most subsets and metrics. 
In contrast, GPT-4o performs better with combinations such as $w/\; A^{++/+}$ and $w/\; A^{++/0}$. 
This highlights that the effects of learning from disagreement samples differ between different LLMs, which is closely related to the ambiguous characteristics of these samples.

\subsection{Fine-tuning with Qwen2.5-72B}\label{sec_ft_qwen}

We replicate the instruction fine-tuning experiment from Section \ref{sec_ft} using Qwen2.5-7B, training with the same instruction data.
Based on the results shown in Table \ref{ft_qwen}, we observe conclusions that are largely consistent with those on LLaMa3-8B.
Specifically, Qwen2.5-7B performs best with medium agreement ($A^+$) when fine-tuned using a single category of annotation agreement across various subsets and metrics.
Compared to unanimous agreement samples ($A^{++}$), training with disagreement samples helps the model’s confidence better align with the degree of human annotation agreement, resulting in a lower MSE. 
When fine-tuning with combinations of different agreement categories, using lower agreement combinations (e.g., $w/\; A^{++/0}$ and $w/\; A^{+/0}$) leads to higher detection accuracy and closer alignment with the annotations, compared to higher agreement combinations ($w/\; A^{++/+}$), especially in the detection of disagreement subsets.
These findings further demonstrate the importance of learning from disagreement samples during instruction fine-tuning, which enhances the model's understanding and judgment of offensive language, particularly for ambiguous samples.

\subsection{In-Depth Analysis of the Generalization Ability of Trained Models}\label{OOD}

In this section, we further explore the generalization ability of the trained models using disagreement samples and conduct an in-depth analysis from a more fundamental perspective of the bias-variance trade-off theory.

In the few-shot experiments (Table \ref{few_shot} and \ref{few-shot_qwen}), we observed a slight accuracy drop when incorporating all three levels of annotation agreement (LLaMa-3 using $A^{++/+/0}$: 83.51\%) compared to using a more balanced combination (LLaMa-3 using $A^{++/0}$: 83.8\%). 
This aligns with the classic trade-off: including overly complex or ambiguous training data may increase variance, leading to overfitting and reduced generalization. Conversely, training only on unanimous samples may lead to underfitting, as the model may not sufficiently learn nuanced decision boundaries.

To further explore this hypothesis, we fine-tuned LLaMA-3 on subsets of MD-Agreement with different agreement levels and evaluated its generalization performance on the Implicit Hate Corpus dataset (IHC) \cite{DBLP:conf/emnlp/ElSheriefZMASCY21}, which contains binary-labeled offensive language samples. 
Since IHC does not include annotator agreement metadata, we focused purely on classification accuracy as an out-of-distribution (OOD) metric. 
The experimental results are presented in Table \ref{OOD-IHC}.

The best OOD performance was achieved when fine-tuning on medium-agreement ($A^+$) samples, suggesting that they strike a favorable balance between challenge and clarity. 
This finding further confirms that annotation disagreement level can act as a proxy for data complexity, and that moderate complexity helps mitigate both bias and variance.

% \begin{table}
%   \centering
%     \begin{tabular}{lrrrr}
%           & \multicolumn{1}{l}{MD (Acc.)} & \multicolumn{1}{l}{MD (MSE)} & \multicolumn{1}{l}{IHC (Acc.)} & \multicolumn{1}{l}{IHC (F1)} \\
%     LLama-3 (zero-shot) & 70.92 & 0.1856 & 58.87 & 47.97 \\
%     $\qquad \qquad w/ \; A^{++}$ & 75.79 & 0.1671 & 59.19 & 48.23 \\
%     $\qquad \qquad w/ \; A^{+}$ & 77.04 & 0.1552 & 59.35 & 48.41 \\
%     $\qquad \qquad w/ \; A^{0}$ & 73.99 & 0.1348 & 59.05 & 48.12 \\
%     \end{tabular}%
%   \caption{Add caption}
%   \label{OOD-IHC}%
% \end{table}%

\begin{table}
\renewcommand{\arraystretch}{1.2}
\small
  \centering
    \begin{tabular}{l|cc|cc}
    \hline
          & \multicolumn{2}{c|}{MD (ID)} & \multicolumn{2}{c}{IHC (OOD)} \\
    \hline
    \multicolumn{1}{c|}{Model} & Acc.$\;\uparrow$  & MSE$\;\downarrow$   & Acc.$\;\uparrow$  & F1$\;\uparrow$ \\
    \hline
    LLaMa3-8B  & 70.92 & 0.1856 & 58.87 & 47.97 \\
    \hline
    $\qquad w/ \; A^{++}$ & 75.79 & 0.1671 & 59.19 & 48.23 \\
    $\qquad w/ \; A^{++}$ & 77.04 & 0.1552 & 59.35 & 48.41 \\
    $\qquad w/ \; A^{++}$ & 73.99 & 0.1348 & 59.05 & 48.12 \\
    \hline
    \end{tabular}%
    \vspace{-0.05in}
  \caption{Performance of fine-tuned LLaMa3-8B on the test set of MD-Agreement and IHC.}
    \vspace{-0.025in}
  \label{OOD-IHC}%
\end{table}%

\begin{table}[t]
\small
  \centering
    \begin{tabular}{lccc}
    \toprule
    Category & \textit{N.}     & \textit{O.}     & \textit{Num} \\
    \midrule
    Swearing & 173   & 415   & 588 \\
    Threatening & 27    & 18    & 45 \\
    Personal Bias & 661   & 402   & 1,063 \\
    Rhetorical Question & 57    & 30    & 87 \\
    Reported Speech & 64    & 27    & 91 \\
    Word Play & 24    & 22    & 46 \\
    Sarcasm & 112   & 50    & 162 \\
    Analogy & 41    & 45    & 86 \\
    False Assertion & 18    & 12    & 30 \\
    Ungrammatical & 10    & 8     & 18 \\
    \midrule
    No context & 258   & 76    & 334 \\
    Not Complete & 6     & 2     & 8 \\
    Noise & 12    & 0     & 12 \\
    \midrule
    Total & 1,463  & 1,107  & 2,570 \\
    \bottomrule
    \end{tabular}%
  \caption{Statistics of the MD-Agreement-v2 \cite{DBLP:conf/eacl/SandriLTJ23}, where \textit{N.} and \textit{O.} represent non-offensive and offensive samples, respectively.}
  \vspace{-0.075in}
  \label{fine-statistics}%
\end{table}%

\subsection{Impact of Linguistic Features in Disagreement Samples}

We evaluate the performance of LLMs on disagreement samples with varying linguistic features in the zero-shot setting.
We use MD-Agreement-v2 \cite{DBLP:conf/eacl/SandriLTJ23} as the benchmark, which is an extended version of MD-Agreement that provides fine-grained annotations of the linguistic features present in each sample, including \textit{Swearing}, \textit{Threatening} language, sensitive topics provoking ambiguity (\textit{Personal Bias}), \textit{Word Play}, \textit{Rhetorical Questions} reflecting the publisher’s viewpoint, \textit{Reported Speech}, \textit{Sarcasm}, \textit{Analogy}, \textit{False Assertions}, or the use of \textit{Ungrammatical} words or sentence structures.
The definitions of features and dataset statistics are provided in Table \ref{fine-statistics} and \ref{disagreement_types}.
Four widely used models are selected for this assessment, including GPT-4, Claude-3.5, Mixtral 8x22B, and LLaMa-3 70B.
The F1-score serves as the evaluation metric. 
Based on the results shown in Figure \ref{dis_type}, the following conclusions can be drawn:

\textbf{LLM exhibits weak performance in detecting samples with complex, context-dependent linguistic features and noise.}
When processing samples containing expressions of \textit{Sarcasm} and \textit{Rhetorical Questions}, LLMs consistently exhibit significantly poor performance, with average F1 scores of 54.6\% and 61.6\%, respectively, which notably lower the overall dataset performance of 67.7\%. 
In contrast, for \textit{Swearing} and \textit{Analogy} samples, which feature explicit token-level markers like swear words or comparative terms, LLMs achieve F1 scores of 83.8\% and 71.2\%, respectively. 
This indicates that samples containing complex linguistic features, which require the model to integrate complete and nuanced context for reasoning and judgment, remain a challenge for existing LLMs. 
Another lower-performing category is the one containing \textit{Ungrammatical} expressions, with an F1 score of only 57.1\%, reflecting the limited ability of LLMs to handle noise, which further reduces their reliability in real-world text moderation tasks.

\textbf{Performance varies significantly across different LLMs, even when detecting the same linguistic feature.}
When detecting samples involving \textit{Reported Speech} that cites others' viewpoints, the best-performing model, LLaMa3-70B, achieved an F1 score 16.3\% higher than the weakest Mixtral-8$\times$22B. 
In addition, for samples containing \textit{Rhetorical Questions}, GPT-4 outperformed Claude-3.5 by 10.3\%, despite similar overall performance across the dataset. 
This underscores the substantial performance differences among models when handling offensive language with the same linguistic feature. 
These differences stem from variations in the models' training mechanisms and the datasets used for training. 
In practical applications, selecting the most suitable model for detecting samples with specific linguistic features may yield better results.

\begin{figure}
\centering
\includegraphics[width=6.75cm]{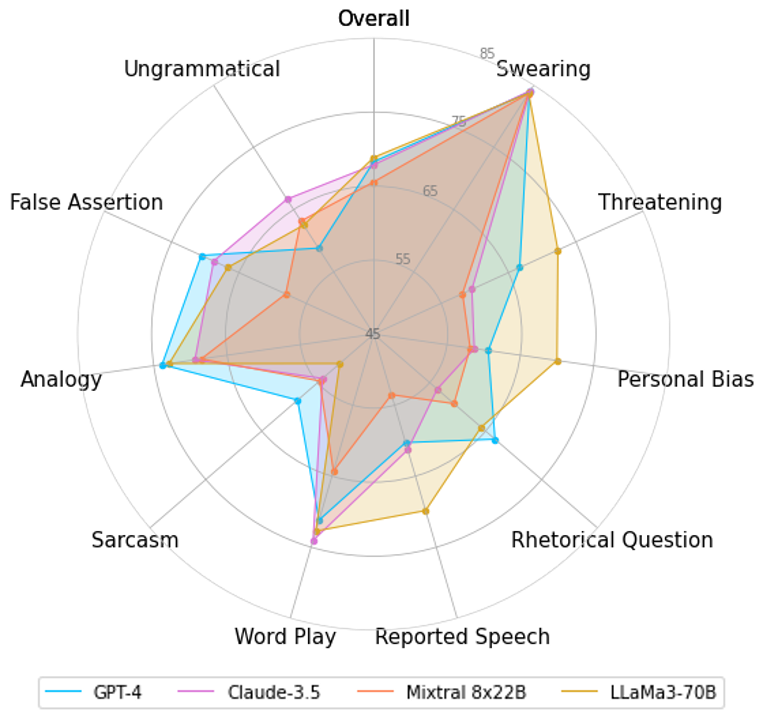}
\vspace{-0.05in}
\caption{Performance of LLMs in detecting disagreement samples with varying linguistic features, where \textit{Overall} refers to the MD-Agreement-v2.
}
\vspace{-0.125in}
\label{dis_type}
\end{figure}

\begin{table*}
\small
\renewcommand{\arraystretch}{1.1}
  \centering
    \begin{tabular}{m{0.75cm}<{\centering}m{10cm}m{1cm}<{\centering}m{1cm}<{\centering}m{1cm}<{\centering}}
    \toprule
    Exp.   & \multicolumn{1}{c}{Sentence} & Label & GPT4o & LLaMa3 \\
    \midrule
    (a)     & Why does the Minneapolis mayor look like a more Soy version of Justin Trudeau & $O^{+}$    & $O^{0}$    & $N^{0}$ \\
    (b)     & This is Amerikkka! \#blacklivesmatter <url> & $N^0$    & $O^{++}$   & $O^{++}$ \\
    (c)     & Hey \#BlackLivesMatter SAY THEIR NAMES! Look these are YOUR PEOPLE! Say Their Names! Do It! Where is Your Outrage Now?! Come forth You \#Hypocrites! Did They Die In Vain?! Ye Generation Of Vipers! These Are Your People Murdered! & $O^{0}$    & $N^{0}$    & $N^{++}$ \\
    (d)     & Doesn't have a kid's cancer charity he can draw from? \#Elections2020 & $N^+$    & $O^{++}$   & $N^{++}$ \\
    \bottomrule
    \end{tabular}%
    \vspace{-0.025in}
  \caption{Error analysis of samples with annotation disagreement in offensive language detection.}
    \vspace{-0.05in}
  \label{case_study}%
\end{table*}%

\begin{table*}[t]
\small
\renewcommand{\arraystretch}{1.05}
  \centering
    \begin{tabular}{m{2cm}m{6cm}m{6.5cm}}
    \toprule
    Taxonomy & \multicolumn{1}{l}{Linguistic Definition} & \multicolumn{1}{l}{Example} \\
    \midrule
    Swearing & The sample includes swearing words or insults (e.g., "prick", "retarded"). & 2nd wave about to be a bitch. ($O^0$)   \\
    \midrule
    Threatening & The sample contains violent expressions or linguistic threats.  & U r going to jail. ($N^0$)  \\
    \midrule
    Personal Bias & The sample includes specific topics that can be interpreted subjectively (e.g., "racist") or expresses specific critical opinions (e.g., "no vax"). & \#DemocratsAreDestroyingAmerica \#Black Lives Matter is a terrorist organization. ($N^0$)  \\
    \midrule
    Rhetorical Question & The sample contains a question asked to highlight a concept, rather than to seek an answer. & Why do we treat our prisoners like this? Is it really because we decided once you commit a crime you’re worthless non-human? ($N^0$)  \\
    \midrule
    Reported Speech & The sample reports something someone else stated (e.g., a newspaper headline). & White Bystanders With Rifles Stare Down George Floyd Protesters: 'You Ain't Got No Guns'. ($O^0$) \\
    \midrule
    Word Play & The sample uses literary devices that alter certain words, including acronyms, alliterations or puns. & The only people ripping this country apart are your fellow liberal \#DemocRATS and your militant concubines. ($N^0$)  \\
    \midrule
    Sarcasm & The sample aims to express the opposite of its surface meaning humorously or to mock someone/something. & Who knew a side effect of COVID would be gross incompetence. ($N^0$)  \\
    \midrule
    Analogy & The sample contains metaphors or analogies. & Trump is a walking petri dish. His goal is to\newline{}spread the virus to as many people as possible. ($N^0$)  \\
    \midrule
    False Assertion & The publisher expresses something contrary to his thoughts, or something false and exaggerated in relation to the context. & Another attempt backfired on them, George Floyd cured Covid-19 and opened up the economy! ($N^0$) \\
    \midrule
    Ungrammatical & The sample includes non-standard varieties, slang, code-switching, or simple typing errors. & Mane it’s hard for some of da blaxk people out dere when dey go into a store they got everybody looking at them "what they doing".  ($N^+$)   \\
    \midrule
    No Context & The sample contains ambiguous pronouns or links, lacking sufficient context. & Dude this guy is serious? And trump retweeted this?? Please anonymous take them out. ($N^+$) \\
    \midrule
    Not Complete & The sample does not convey complete information or miss some parts.  & Wtaf, a farce in three parts: ($O^0$) \\
    \midrule
    Noise & The sample is clearly not offensive but marked as such (or vice versa). & In a singular voice, art across the world. ($N^+$) \\
    \bottomrule
    \end{tabular}%
    % \vspace{-0.05in}
    \caption{Definitions and examples of various linguistics in the offensive language with annotation disagreement, with definitions from \cite{DBLP:conf/eacl/SandriLTJ23} and examples from the MD-Agreement-v2 dataset.
    Three categories - \textit{No Context}, \textit{Not Complete}, and \textit{Noise} - arise from the low quality of the data itself and are excluded from LLM evaluation.
    In contrast, we retain evaluation for the \textit{Ungrammatical} category, as it reflects non-standard expressions and typos commonly found in offensive language.
    }
  % \vspace{-0.05in}
  \label{disagreement_types}%
\end{table*}%

\subsection{Error Analysis}

To gain deeper insight into the challenge posed by offensive language with disagreement, we manually inspect the set of samples misclassified by the models and conduct an error analysis. 
The following two main types of errors are identified, with samples and predictions from GPT-4o and LLaMa3-72B shown in Table \ref{case_study} for illustration:

\textbf{Type I error} refers to samples that are labeled as \textit{non-offensive} but are detected as \textit{offensive}.
This error primarily arises from subtle linguistic features such as sarcasm and metaphor, which make the judgment of the sample ambiguous.
For instance, in Example (a), the term “\textit{Amerikkka}” is a variant of “\textit{America}” used to intensify emotional expression. 
Due to insufficient context, most annotators do not consider it offensive. 
However, GPT-4o and LLaMa3, due to their sensitivity to the hashtag \textit{\#blacklivesmatter}, consistently classify it as offensive language.
Similarly, in Example (b), a sarcastic rhetorical question leads to a misclassification by GPT-4o.
This phenomenon highlights the complexity that human annotators face in determining offensive language and also reveals the issue of over-sensitivity in existing LLMs to certain linguistic expressions, resulting in decisions that do not align with human standards.
In future work, we will perform a more detailed analysis of expressions in samples with disagreement annotation and explore how different types of expressions affect model detection performance.

\textbf{Type II error} refers to sentences labeled as offensive but classified as non-offensive by the models.
This error primarily arises from the models lacking or failing to effectively integrate the necessary background knowledge for detecting offensive content, leading to an inaccurate understanding of the sample's true meaning.
For example, in Example (c), the comparison between \textit{the mayor of Minneapolis} and \textit{Justin Trudeau} uses “\textit{Soy}” as an adjective, which implies weakness and is intended to belittle the mayor. 
Both human annotators and GPT-4o capture the offensive nature of the sample, but LLaMa3 fails to correctly identify its offensiveness due to insufficient relevant knowledge.
In Example (d), the phrase “\textit{Ye Generation of Vipers}”, a religiously charged expression, is used to strongly criticize police brutality against black people. However, the model fails to integrate the context, leading to a missed detection.
We plan to introduce more comprehensive background knowledge to enhance the understanding capability of LLMs and explore the performance of knowledge-enhanced models in detecting disagreement samples.

% \textbf{Type I error} refers to sentences labeled as \textit{offensive} but classified as \textit{non-offensive} by the models.

% \textbf{Type II error} refers to instances labeled as \textit{non-offensive} but detected as \textit{offensive}.

\end{document}